\documentclass{article}

     \PassOptionsToPackage{compress}{natbib}
\usepackage[final]{neurips_2023}

\usepackage[utf8]{inputenc} % allow utf-8 input
\usepackage[T1]{fontenc}    % use 8-bit T1 fonts
\usepackage{hyperref}       % hyperlinks
\usepackage{url}            % simple URL typesetting
\usepackage{booktabs}       % professional-quality tables
\usepackage{amsfonts}       % blackboard math symbols
\usepackage{nicefrac}       % compact symbols for 1/2, etc.
\usepackage{microtype}      % microtypography
\usepackage{xcolor}         % colors

\usepackage{amsmath,amsfonts,bm}

\def\eqref#1{equation~\ref{#1}}
\def\1{\bm{1}}

\def\rva{{\mathbf{a}}}

\def\rvu{{\mathbf{i}}}

\def\rvr{{\mathbf{r}}}

\def\rvu{{\mathbf{u}}}
\def\rvv{{\mathbf{v}}}

\def\rvx{{\mathbf{x}}}
\def\rvy{{\mathbf{y}}}

\DeclareMathAlphabet{\mathsfit}{\encodingdefault}{\sfdefault}{m}{sl}
\SetMathAlphabet{\mathsfit}{bold}{\encodingdefault}{\sfdefault}{bx}{n}

\newcommand{\E}{\mathbb{E}}

\newcommand{\R}{\mathbb{R}}

\newcommand{\KL}{D_{\mathrm{KL}}}

\usepackage{tikz}
\usepackage{subcaption}
\usepackage{graphicx}
\usepackage{algorithm}
\usepackage{algorithmic}
\usepackage{multirow}
\usepackage{wrapfig}
\usepackage[export]{adjustbox}

\usepackage{soul} % for st command

\definecolor{BrickRed}{rgb}{0.6,0,0}
\definecolor{RoyalBlue}{rgb}{0,0,0.8}
\definecolor{Tdgreen}{rgb}{0,0.4,0.7}
\definecolor{cadmiumgreen}{rgb}{0.0, 0.42, 0.24}
\hypersetup{colorlinks, citecolor=RoyalBlue, linkcolor=RoyalBlue}

\title{Hyperbolic VAE via Latent Gaussian Distributions}

\author{%
  Seunghyuk Cho\\
  POSTECH GSAI\\
  \texttt{shhj1998@postech.ac.kr} \\
  \And
  Juyong Lee\\
  KAIST AI\\
  \texttt{agi.is@kaist.ac.kr} \\
  \And
  Dongwoo Kim\\
  POSTECH GSAI \& CSED\\
  \texttt{dongwoo.kim@postech.ac.kr} \\
}

\begin{document}

\maketitle

\begin{abstract}
We propose a Gaussian manifold variational auto-encoder (GM-VAE) whose latent space consists of a set of Gaussian distributions. It is known that the set of the univariate Gaussian distributions with the Fisher information metric form a hyperbolic space, which we call a Gaussian manifold. To learn the VAE endowed with the Gaussian manifolds, we propose a pseudo-Gaussian manifold normal distribution based on the Kullback-Leibler divergence, a local approximation of the squared Fisher-Rao distance, to define a density over the latent space. We demonstrate the efficacy of GM-VAE on two different tasks: density estimation of image datasets and state representation learning for model-based reinforcement learning. GM-VAE outperforms the other variants of hyperbolic- and Euclidean-VAEs on density estimation tasks and shows competitive performance in model-based reinforcement learning. We observe that our model provides strong numerical stability, addressing a common limitation reported in previous hyperbolic-VAEs.
The implementation is available at \url{https://github.com/ml-postech/GM-VAE}.
\end{abstract}
\section{Introduction}

The geometry of latent space in generative models, such as variational auto-encoders (VAE)~\citep{vae}, reflects the structure of the data representations.
\citet{mathieu19,nagano19,rown} show that employing hyperbolic space as the latent space improves the preservation of the hierarchical structure within the data.
The theoretical background for adopting hyperbolic space lies in the analysis of \citet{treeembedding}; the tree-structured data can be embedded with arbitrary low distortion in hyperbolic space, while Euclidean space requires extensive dimensions.
% The expanded geometry is not just limited to the hyperbolic space, as the space can be other types of Riemannian manifolds, such as spherical manifolds~\citep{sphericalVAE1,sphericalVAE2} or the product of Riemannian manifolds with mixed curvatures~\citep{MVAE}.

Previously proposed hyperbolic VAEs rely on 
Poincar\'e normal distribution~\citep{mathieu19} or hyperbolic wrapped normal distribution~\cite{nagano19} for the prior and variational distributions. 
Unlike the Gaussian distribution in Euclidean space, however, these distributions suffer from several shortcomings, including the absence of closed-form Kullback-Leibler (KL) divergence, numerical instability~\citep{mathieu19,MVAE}, and high computational cost in sampling~\citep{mathieu19}.
%So far, the latent space of the hyperbolic VAE has been analyzed from the hyperbolic geometry.
%Through the transformation from the hyperbolic space to the statistical manifold, we can have alternative tools to analyze its latent space.

% \begin{wrapfigure}{r}{0.5\textwidth}
\begin{figure}[t!]
    \centering
    %\vskip 0.1in
    \begin{subfigure}[t]{.4\linewidth}
        \centering
        \begin{tikzpicture}
            \node[anchor=south west] at (0, 0) {\includegraphics[width=\linewidth]{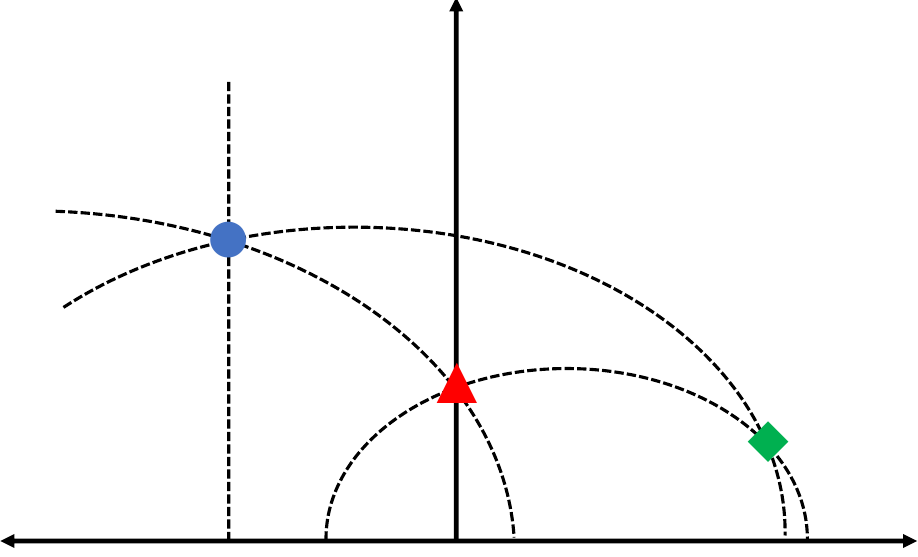}};
            \node[anchor=south west] at (1.5, 2.1) {\tiny $(\mu_1, \sigma_1)$};
            \node[anchor=south west] at (3, 0.8) {\tiny $(\mu_2, \sigma_2)$};
            \node[anchor=south west] at (4.65, 0.85) {\tiny $(\mu_3, \sigma_3)$};
            \node[anchor=north west] at (5.3, 0.25) {\tiny $\mu$};
            \node[anchor=south west] at (2.85, 3.2) {\tiny $\sigma$};
        \end{tikzpicture}
        \caption{Gaussian manifold}
    \end{subfigure}
    \begin{subfigure}[t]{.4\linewidth}
        \centering
        \begin{tikzpicture}
            \node[anchor=south west] at (0, 0.05) {\includegraphics[width=.85\linewidth]{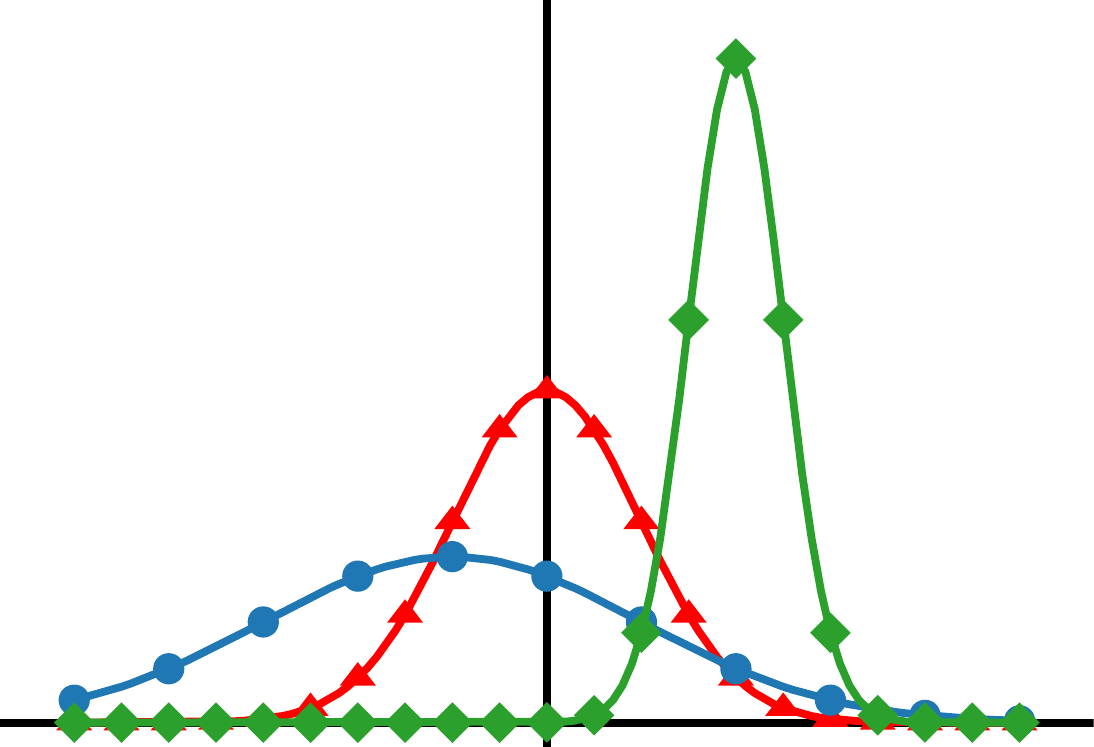}};
            \node[anchor=north west] at (4.5, 0.25) {\tiny $x$};
            \node[anchor=south west] at (2.5, 3.2) {\tiny $p(x)$};
        \end{tikzpicture}
        \caption{Univariate Gaussians}
    \end{subfigure}
    \caption{
    (a) The visualization of the Gaussian manifold consisting of a set of Gaussian distributions. Each point of the Gaussian manifold is a pair of two parameters of a univariate Gaussian distribution: ($\mu$, $\sigma$) $\in \mathbb{R}\times\mathbb{R}_{>0}$. The dashed lines are the geodesics, which are either the ellipses with eccentricity $1/\sqrt{2}$ with the origin placed on the $\mu$-axis or straight lines parallel to the $\sigma$-axis. 
    (b) Three univariate Gaussian distributions correspond to three points in the Gaussian manifold in (a).
    }
    \label{fig:gaussian_manifold}
\end{figure}
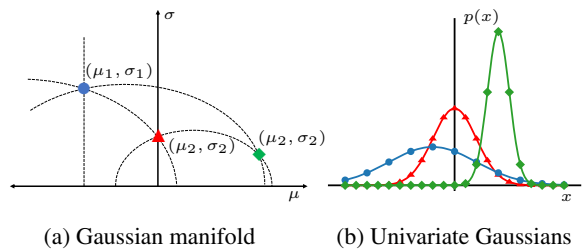
% \end{wrapfigure}

Meanwhile, we can form a Riemannian manifold from the set of univariate Gaussian distributions by equipping the Fisher information metric (FIM). %; we remind that the Riemannian manifold is a pair of a manifold and a metric tensor, where the metric tensor induces the shortest distance between two points which is called the geodesic distance.
% The Riemannian manifold shares its manifold with the Poincar\'e half-plane model which is one of the four isometric hyperbolic models.
It is known that the FIM of univariate Gaussian distributions is akin to that of the metric tensor of the Poincar\'e half-plane model~\citep{gmhyperbolic}, providing a perspective of viewing the points in hyperbolic space as univariate Gaussian distributions. In other words, a Gaussian distribution can be mapped to a single point in the open half-plane manifold as shown in \autoref{fig:gaussian_manifold}, where the FIM forms the shortest geodesic distance between two Gaussian distributions.
% This statistical manifold is known to have a metric tensor akin to that of the Poincar\'e half-plane~\citep{gmhyperbolic}, providing a perspective of viewing the points of hyperbolic space as univariate Gaussian distributions.
%Since all hyperbolic models having the same value of curvature are isometric to each other, the statistical manifold of the univariate Gaussian distribution is isometric to a hyperbolic space. 
Noting that the numerical issue of Poincar\'e normal arises from the geodesic distance of hyperbolic space, we question whether this perspective can lead us to define a new distribution with better analytic properties.%, especially a closed-form KL divergence between the distributions.%, as well as a corresponding variant of VAE.

% \jylee{
In this work, inspired by the fact that KL divergence itself is a statistical distance that locally approximates the geodesic distance~\citep{pglove}, we propose a hyperbolic distribution by substituting the geodesic distance of Poincar\'e normal with the KL divergence between the univariate Gaussian distributions.
We then verify that this simple yet powerful alteration results in several practical analytic properties; the proposed distribution reduces into the product of two well-known distributions, i.e., the Gaussian and gamma distributions, which are easy to sample, with a closed-form KL divergence between the proposed distributions.
% we verify two properties: the tractability of KL divergence between the Riemannian normal distributions defined with KL divergence as the distance function and the generalizability of the local approximation property toward the hyperbolic space of arbitrary curvature.
By adopting the proposed hyperbolic distribution, we introduce a new variant of hyperbolic VAE, named Gaussian manifold VAE (GM-VAE), whose latent space is a set of Gaussian distributions. %Gaussian manifold.

During the experiments, we observe that the proposed distribution is robust in terms of sampling and KL divergence computation compared to the commonly-used hyperbolic distributions; we briefly explain the reason why others are numerically unstable. 
Experimental results on the density estimation task with image datasets show that GM-VAE can achieve outperforming generalization performances to unseen data against baselines of Euclidean and hyperbolic VAEs. 
Application of GM-VAE on model-based reinforcement learning (RL) verifies the feasibility of using hyperbolic space on another domain of task.
% Analysis of the latent space exhibits that the geometrical structures and probabilistic semantics of the dataset can be captured in the representations learned with GM-VAE.
%Notable properties of GM-VAE, including numerical stability and behaviors in representation learning, are also present in several analyses.

We summarize our contributions as follows:
\begin{itemize}
    \item We introduce a variant of VAE whose latent space is defined on a statistical manifold formed by univariate Gaussian distributions, namely Gaussian manifold.
    \item We propose a new distribution, called a pseudo Gaussian manifold normal distribution, which is easy to sample and has closed-form KL divergence, to train VAE on the Gaussian manifold.
    % \item We propose new encoder and decoder structures to support the proper transition between Euclidean (data) space and the statistical manifold.
    \item We empirically verify that the newly proposed VAE performs stable training without numerical issues on the density estimation task with several image datasets. The proposed model outperforms the baseline Euclidean VAE and other hyperbolic variants.
    \item We show that our method can be used for model-based RL. Specifically, we replace the latent space of the world model with hyperbolic space for learning environments, showing competitive results with a state-of-the-art baseline.
\end{itemize}

\section{Preliminaries}
In this section, we first review the fundamental concepts of hyperbolic space and commonly used hyperbolic models. We then explain the Riemannian geometry between statistical objects, showing the connection between the statistical manifold and hyperbolic space.

\subsection{Review on hyperbolic space}
\label{sec:riemannian_manifold}

% Riemannian manifold
\paragraph{Riemannian manifold.}
A $n$-dimensional Riemannian manifold consists of a manifold $\mathcal{M}$ and a metric tensor $g: \mathcal{M} \rightarrow \R^{n \times n}$, which is a smooth map from each point $\rvx \in \mathcal{M}$ to a symmetric positive definite matrix.
The metric tensor $g(\rvx)$ defines the inner product of two tangent vectors for each point of the manifold $\langle \cdot, \cdot \rangle_{\rvx}: \mathcal{T}_{\rvx}\mathcal{M} \times \mathcal{T}_{\rvx}\mathcal{M} \rightarrow \R$, where $\mathcal{T}_{\rvx}\mathcal{M}$ is the tangent space of $\rvx$.

The metric tensor induces basic Riemannian operations, such as a geodesic, exponential map, log map, and parallel transport.
Given two points $\rvx, \rvy \in \mathcal{M}$,  geodesic $\gamma_{\rvx}: [0, 1] \rightarrow \mathcal{M}$ is a unit speed curve on $\mathcal{M}$ being the shortest path between $\gamma(0) = \rvx$ and $\gamma(1) = \rvy$. 
This curve can be interpreted as a generalized path of a straight line in Euclidean space.
The exponential map $\exp_{\rvx}: \mathcal{T}_{\rvx}\mathcal{M} \rightarrow \mathcal{M}$ is defined as $\exp_{\rvx} (\rvv) = \gamma(1) = \rvy$ given $\gamma$ is a geodesic starting from $\gamma(0) = \rvx$ and $\gamma'(0) = \rvv \in \mathcal{T}_{\rvx}\mathcal{M}$.
The log map $\log_{\rvx}: \mathcal{M} \rightarrow \mathcal{T}_{\rvx}\mathcal{M}$ is the inverse of the exponential map, i.e., $\log_{\rvx}(\exp_{\rvx}(\rvv)) = \rvv$.
The parallel transport $\textrm{PT}_{\rvx \rightarrow \rvy}: \mathcal{T}_{\rvx}\mathcal{M} \rightarrow \mathcal{T}_{\rvy}\mathcal{M}$ moves the tangent vector $\rvv$ along the geodesic between $\rvx$ and $\rvy$.
The geodesic distance $d_{\mathcal{M}}(\rvx, \rvy)$ can be induced by the metric tensor as follows:
\begin{equation*}
    d_{\mathcal{M}}(\rvx, \rvy) = \int_{0}^{1} \sqrt{ \langle \dot{\gamma}(t),\dot{\gamma}(t) \rangle_{\gamma(t)} } dt.
\end{equation*}

\paragraph{Hyperbolic space.}
% curvature
One method of classifying Riemannian manifolds, a basic question of differential geometry, is based on curvature.
Among different types of curvatures, one popular curvature is the sectional curvature $\kappa_g$, which is a generalization of the Gaussian curvature in classical surface geometry. 
Given two linearly independent vector fields $X, Y \in \mathfrak{X}(\mathcal{M})$,
the sectional curvature $\kappa_g$ can be computed with Riemannian curvature tensor $R \colon {\mathfrak {X}}(\mathcal{M})\times {\mathfrak {X}}(\mathcal{M})\times {\mathfrak {X}}(\mathcal{M})\rightarrow {\mathfrak {X}}(\mathcal{M})$ as below:
\begin{align*}
\kappa_g = \frac{\langle R(X,Y)Y, X \rangle}{\langle X, X \rangle \langle Y, Y \rangle - \langle X, Y \rangle^2},
\end{align*}
where $R(X,Y)Y$ returns a tensor field assigning a tensor to each point of the Riemannian manifold $\mathcal{M}$.
%\dw{in the above description $R$ needs three arguments whereas $R$ in the equation takes two}
The hyperbolic space is a Riemannian manifold that has the sectional curvature value of constant negative~\citep{lorentz}. 
The hyperbolic space is known to be able to embed tree-structured data with arbitrarily low distortion~\citep{treeembedding}.

\paragraph{Hyperbolic models.}
We utilize three famous models of hyperbolic space: the Poincar\'e disk model, the Lorentz model, and the Poincar\'e half-plane model.

The Poincar\'e disk model is a hyperbolic space with an open disk manifold.
Earlier hyperbolic machine learning work uses the Poincar\'e disk model because it has a simple closed-form of the operations, such as exponential and log maps~\citep{mathieu19}.
However, the Poincar\'e disk model suffers from numerical stability issues when the points exist near the boundary of the manifold.

The Lorentz model is often used as an alteration of the Poincar\'e disk model~\citep{lorentz,nagano19,hyperbolic_normalizing_flow,rown}.
The Lorentz model uses a half hyperboloid manifold, where the closed form of the Riemannian operations exists, so the numerical stability issue of employing the Poincar\'e disk model is relieved.

The Poincar\'e half-plane model is another well-known model of hyperbolic space with an open half-plane manifold. 
The metric tensor of a point of the two-dimensional Poincar\'e half-plane model $(x, y)$ is $y^{-2}\textrm{diag}(1, 1)$.

\paragraph{Numerical stability issues of the hyperbolic models.}
Hyperbolic space suffers from numerical stability when applied to machine learning algorithms~\citep{machineprecisionerror1,MVAE,mathieu19}.
The numerical stability mainly occurs for two reasons: machine precision error and unstable Riemannian operations.

First, due to the machine precision error, the hyperbolic points represented with floating point differ from the real value~\citep{machineprecisionerror1,machineprecisionerror2}. 
In contrast to Euclidean space, the points of hyperbolic space need to satisfy manifold constraints, e.g., the Poincar\'e disk model allows points whose Euclidean norm is less than one.
A point can be placed near the boundary during the optimization or inference processes. Although the point does not violate the manifold constraint in theory, it can be located on or out of the boundary when represented with a floating point due to machine precision error.
%When a hyperbolic point is near the boundary of the inequality constraint or meets the equality condition, if the hyperbolic point contain values which occurs machine precision error in floating point representation, the hyperbolic point escapes from the manifold when converted to floating point.
We empirically observe that the manifold constraint violation occurs frequently when we need to embed many data points in hyperbolic space. % where embeddings near the boundary condition exist and contain large values which especially have high machine precision error.
% e.g., when we use the Poincar\'e disk model, the points can be located out of the unit disk.
\autoref{fig:machine_precision_error} demonstrates the machine precision error of each hyperbolic model.

Second, the Riemannian operations of hyperbolic space can result in a not-a-number (NaN) value when the input value is not in the manifold.
For example, the geodesic distance from the Poincar\'e disk model and the log mapping of the Lorentz model are unstable Riemannian operations, which are written as:
% \begin{align}
%     d_{\mathcal{P}}(\rvx, \rvy) &= \cosh^{-1}\left( 1 + 2\frac{\Vert \rvx - \rvy \Vert^2}{(1-\Vert \rvx \Vert^2)(1-\Vert \rvy \Vert^2)} \right)\\
%     \log_{\rvu}(\rvv) &= \frac{\cosh^{-1}(\alpha)}{\sqrt{\alpha^2-1}}(\rvv - \alpha \rvu),
% \end{align}

\begin{equation*}
    d_{\mathcal{P}}(\rvx, \rvy) = \cosh^{-1}\left( 1 + 2\frac{\Vert \rvx - \rvy \Vert^2}{(1-\Vert \rvx \Vert^2)(1-\Vert \rvy \Vert^2)} \right), \, \log_{\rvu}(\rvv) = \frac{\cosh^{-1}(\alpha)}{\sqrt{\alpha^2-1}}(\rvv - \alpha \rvu),
\end{equation*}

where $\alpha = \rvu_0\rvv_0 - \sum_{i=1}^n\rvu_i\rvv_i$ is the Lorentzian inner product of $\rvu, \rvv$. 
For the Poincar\'e disk model, when the points $\rvx, \rvy$ are near the boundary of the unit disk, the floating point representation of the norm values $\Vert \rvx \Vert^2, \Vert \rvy \Vert^2$ becomes one. The denominator of the geodesic distance then becomes zero.
% \dw{In which situation is the norm greater than or equal to one? In other words, why can one not enforce the constraints through the algorithmic perspective? Is this the consequence of the machine precision error? e.g., 0.999999 $\rightarrow$ 1?}
For the Lorentz model, if $\rvu = \rvv$ and $\rvu$ contains large values in the coordinates, $\alpha$ becomes less than one which results in NaN because the domain of $\cosh^{-1}(x)$ is $x \geq 1$.

\subsection{Statistical manifold of univariate Gaussians}
\label{sec:statistical_manifold}
A particular case of the Riemannian manifold is a statistical manifold, where each point in the manifold corresponds to a probability distribution.
Specifically, the parameter manifold $\mathcal{M}$ of the probability distributions $p_\theta:\mathcal{X} \rightarrow \mathbb{R}$, where $\theta \in \mathcal{M}$, equipped with the Fisher information metric (FIM) forms a Riemannian manifold~\citep{fim}. The FIM is defined as:
\begin{equation*}
    g_{ij}(\boldsymbol{\theta}) = \int_{\mathcal{X}} \frac{\partial \log p_\theta(x)}{\partial \theta_j} \frac{\partial \log p_\theta(x)}{\partial \theta_j} p_\theta(x) \, dx.
\end{equation*}

In the parameter space of univariate Gaussian distributions $\{(\mu, \sigma) \mid \mu \in \R, \sigma \in \R_{>0}\}$, the FIM can be simplified as two-dimensional diagonal matrix $\sigma^{-2}\textrm{diag}(1, 2)$~\citep{gmhyperbolic}.

\paragraph{Connection to the Poincar\'e half-plane model.}
% Hyperbolic half-plane
The diagonal form of the FIM implies that the Riemannian manifold with  $\{(\mu, \sigma) \mid \mu \in \R, \sigma \in \R_{>0}\}$ has the same set of points as the manifold of the Poincar\'e half-plane, but with different curvature value of $-0.5$.

The parameter space of the $n$-dimensional diagonal Gaussian distributions becomes the product of $n$ manifolds of the parameter space of univariate Gaussian distributions.
In turn, the statistical manifold of $n$-dimensional diagonal Gaussian distributions can be viewed as the product of $n$ hyperbolic spaces.
%\jylee{Don't we need to explain any explicit connection of this product space with hyperbolic space?}
The operations on the product of the Riemannian manifolds $\bigotimes_{i=1}^n \mathcal{M}_i$ are defined manifold-wise.
For example, an exponential map applied on a point $(p_i)_{i=1}^n \in \bigotimes_{i=1}^n \mathcal{M}_i$, with tangent vector $v_i \in \mathcal{T}_{p_i}\mathcal{M}_i$ for each $i \in \{1, \cdots, n \}$, can be represented as $(\exp_{p_i}(v_i))_{i=1}^n $.

\paragraph{Distance in the statistical manifold.}
In the statistical manifold, distance functions measure the difference between two distributions on the statistical manifold.
One example is the geodesic distance derived from the FIM, which is called the Fisher-Rao distance.
The Fisher-Rao distance of the statistical manifold of univariate Gaussian distributions is the same as the geodesic distance of the Poincar\'e half-plane model with constant negative curvature $-0.5$.

KL divergence is another widely-used statistical distance for distributions, defined as $\KL(p \parallel q) := \int_x p(x) \log \frac{p(x)}{q(x)} \, dx$ for two distributions $p, q$ in the same statistical manifold.
One notable property of KL divergence is that it can locally approximate the squared Fisher-Rao distance~\citep{pglove}:
\begin{align*}
    \KL(p(\cdot; \boldsymbol{\theta} + d\boldsymbol{\theta}) \parallel p(\cdot; \boldsymbol{\theta}))
    = \frac{1}{2}\sum_{ij} g_{ij}(\boldsymbol{\theta}) d\theta_i d\theta_j + \mathcal{O}(\Vert d\boldsymbol{\theta}\Vert^3).
\end{align*}
\section{Method}
In this section, we first present the concept of the Gaussian manifold, which can have an arbitrary curvature by reparameterizing univariate Gaussian distribution. We then propose a pseudo Gaussian manifold normal distribution. Finally, we suggest a new variant of the VAE defined over the Gaussian manifold with PGM normal as prior.
We denote the density function of the Gaussian distribution as $\mathcal{N}(x; \mu, \sigma^2) = 1/(\sqrt{2\pi\sigma^2}) \exp\left( -(\mu - x)^2/(2\sigma^2) \right)$.

\subsection{Gaussian manifold with arbitrary curvature}
Previous studies on hyperbolic space emphasize the importance of having an arbitrary curvature~\citep{MVAE, mathieu19}. These works empirically show that the generalization performances of hyperbolic VAEs can be improved with varying curvatures. However, as shown in \autoref{sec:statistical_manifold}, the univariate Gaussian distributions form a manifold with curvature value of $-0.5$, limiting the flexibility of the manifold.
%Before using the manifold as the latent space of VAE, we propose an extension of this statistical manifold to incorporate an arbitrary curvature $-c$, as the efficacy of hyperbolic space with an arbitrary curvature has been reported~\citep{MVAE, mathieu19}.

We show that the statistical manifold of univariate Gaussian distributions can have an arbitrary curvature by reparameterizing the univariate Gaussian distribution properly.
Let $\mathcal{N}(\sqrt{2c}\mu, \sigma^2)$ be the reparameterized univariate Gaussian distribution with additional parameter $c > 0$. The reparameterization leads to the FIM of $\sigma^{-2}\textrm{diag}(1, 1/c)$ showing that the curvature of the statistical manifold is $-c$. 
The computation of the sectional curvature of the extended FIM is described in Appendix~\ref{apx:fisher_extension}.

We call the statistical manifold with the reparameterized univariate Gaussian distributions and the extended FIM as the Gaussian manifold and denote it as $\mathcal{G}_c$, where $-c$ is the curvature of the Gaussian manifold.
%This extension can be performed by adopting the extended FIM which can be written as $\sigma^{-2}\textrm{diag}(1, c)$, where $c \in \mathbb{R_{>0}}$. 
%The detailed transformation from FIM to the extended FIM is described in Appendix~\ref{apx:fisher_extension}.

We then verify that the KL divergence between the distributions of the Gaussian manifold approximates the geodesic distance, even in the presence of arbitrary curvature in the Gaussian manifold. Let $(\mu, \sigma) \in \mathcal{G}_c$ be an arbitrary point of the Gaussian manifold. The KL divergence between $(\mu, \sigma)$ and its neighbor $(\mu + d\mu, \sigma + d\sigma)$ can be computed as:
\begin{align}
    \label{eq:kl_gm}
    \frac{\KL(\mathcal{N}(\sqrt{2c}(\mu + d\mu), (\sigma + d\sigma)^2) || \mathcal{N}(\sqrt{2c}\mu, \sigma^2))}{2c}
    = \frac{1}{2}\begin{pmatrix}d\mu \\d\sigma \\\end{pmatrix}^{\hspace{-.5em}\top} \hspace{-.4em} \begin{pmatrix}\frac{1}{\sigma^2} & 0 \\ 0 & \frac{1}{c\sigma^2}\\\end{pmatrix}  \begin{pmatrix}d\mu \\d\sigma \\\end{pmatrix} + \mathcal{O}((d\sigma)^3),
\end{align}
where the first term is the squared Riemannian norm of the tangent vector $(d\mu, d\sigma)$ approximating the squared Fisher-Rao distance.
The detailed derivation of the KL divergence of the Gaussian manifold is described in Appendix~\ref{apx:kl_gm}.
%The details of the derivations are available in Appendix~\ref{apx:kl_gm}.

\subsection{Pseudo Gaussian manifold normal distribution}
We propose a pseudo Gaussian manifold (PGM) normal distribution defined over the Gaussian manifold. 
Let $(\mu, \sigma) \in \mathcal{G}_c$ be a point in the Gaussian manifold.
Inspired by the Riemannian normal distribution~\citep{riemanniannormal}, we define the probability density function of PGM normal with the KL divergence as:
\begin{align} 
\label{eq:PGMNormal}
    \mathcal{K}_c(\mu, \sigma; \alpha, \beta, \gamma) 
    = \frac{\sigma^3}{Z(c, \beta, \gamma)}
        \times \exp\left(-\frac{\KL(\mathcal{N}(\sqrt{2c} \cdot \mu, \sigma^2) \parallel \mathcal{N}(\sqrt{2c} \cdot \alpha, \beta^2))}{2c \cdot \gamma^2}\right),
\end{align}
where $(\alpha, \beta) \in \mathcal{G}_c$, and $\gamma \in \R_{>0}$ are the parameters of the distribution. The distribution is centered at $(\alpha,\beta)$ with additional scale parameter $\gamma$.
We verify the convergence of the PGM normal and compute the normalizing constant $Z(c, \beta, \gamma)$ over the probability measure of the Gaussian manifold at Appendix \ref{apx:normalizing_constant_details}.
As shown in \autoref{eq:kl_gm}, the KL divergence of the Gaussian manifold approximates the Fisher-Rao distance between $\mathcal{N}(\sqrt{2c} \cdot \alpha, \beta^2)$ and $\mathcal{N}(\sqrt{2c} \cdot \mu, \sigma^2)$. 
Therefore, the PGM normal accounts for the geometric structure of the univariate Gaussian distributions.

The factorization of the probability density function in \autoref{eq:PGMNormal} multiplied with the square root of the determinant of the FIM shows the advantages of the PGM normal, which can be written as:
\begin{align}
\label{eq:factorization}
    \mathcal{K}_c(\mu, \sigma; \alpha, \beta, \gamma) \cdot \sqrt{\operatorname{det}(g)}
    = \mathcal{N}(\mu; \alpha, \beta^2\gamma^2) \cdot 2\sigma \operatorname{Gamma}\left(\sigma^2; \frac{1}{4c\gamma^2} + 1, \frac{1}{4c\beta^2\gamma^2}\right),
\end{align}
where $\operatorname{Gamma}(z; a, b) = \frac{b^a}{\Gamma(a)}z^{a-1}\exp\left(-bz\right)$ and $g$ is the FIM of the Gaussian manifold. 
The $\sqrt{\det g}$ term enables us to sample and compute the KL divergence in an Euclidean manner.
Thanks to the properties of Gaussian and gamma distributions, the PGM normal is easy to sample and has a closed-form KL divergence. 
The detailed derivation is available in Appendix \ref{apx:sampling_details} and Appendix \ref{apx:kl_details}.
The factorization has the same form as the well-known conjugate prior to the Gaussian distribution.
In that sense, the PGM normal explicitly incorporates the geometric structure between Gaussians into the known prior distribution.

We note that the PGM normal can be easily extended for the diagonal Gaussian manifold, a manifold formed by diagonal Gaussian distributions since the diagonal Gaussian manifold is the product of the Gaussian manifolds.

\begin{algorithm}[b!]
% \setstretch{1.3}
\caption{Decoder}
%\hspace*{\algorithmicindent} 
 \textbf{Input}  Parameter $(\alpha, \beta) \in \mathcal{G}_c, \gamma$, Decoding layers $\textrm{Dec}(\cdot)$ \\
 \textbf{Output} Reconstruction $\rvx'$
\begin{algorithmic}[1]
\STATE Sample $\mu \sim \mathcal{N}(\alpha, \beta^2\gamma^2)$
\STATE Sample $\log \sigma^2 \sim \mathrm{Gamma}\left(\frac{1}{4c\gamma^2} + 1, \frac{1}{4c\beta^2\gamma^2}\right)$
%\STATE $\rvz = [\mu, \sigma]$
\STATE $\rvx' = \textrm{Dec}([\mu, \log \sigma^2])$
% \STATE ${\rvv} = T_c^{-1}({\mu}, {\sigma}) $ \COMMENT{$ ({\mu}, {\sigma}) \in \mathcal{G}_c, {\rvv} \in \mathcal{L}^2_c$}
% \STATE $\tilde{\rvv} = \log_{\boldsymbol{0}_\mathcal{L}}^c({\rvv}) $ \COMMENT{$\tilde{\rvv} \in \mathcal{E}$}
% \STATE $\rvx' = \textrm{Dec}(\rvv)$ 
\STATE \textbf{return} $\rvx'$
\end{algorithmic}
\label{alg:dec}
\end{algorithm}

\subsection{Gaussian manifold VAE}
We introduce a Gaussian manifold VAE (GM-VAE) whose latent space is defined over the diagonal Gaussian manifold with the help of the PGM normal. 
We use the PGM normal for variational and prior distributions. 
To be specific, with the PGM normal, the evidence lower bound (ELBO) of the GM-VAE can be formalized with the diagonal Gaussian manifold $\{(\boldsymbol{\mu}, \Sigma) \mid \boldsymbol{\mu} \in \R^n, \Sigma \in \R^n_{>0} \}$ as:
\begin{align*}
    \E_{q_\phi(\boldsymbol{\mu}, \Sigma \mid \rvx) \cdot \sqrt{\det(g)}} \left[ \log p_\theta(\rvx \mid \boldsymbol{\mu}, \Sigma) \right]
    - \KL\left( q_\phi(\boldsymbol{\mu}, \Sigma \mid \rvx) \cdot \sqrt{\det(g)} \parallel p(\boldsymbol{\mu}, \Sigma) \cdot \sqrt{\det(g)}\right),
    \label{eq:elbo}
\end{align*}
where $p_\theta(\rvx \mid \boldsymbol{\mu}, \Sigma)$ is the decoder network, $q_\phi(\boldsymbol{\mu}, \Sigma \mid \rvx)$ is the encoder network and $p(\boldsymbol{\mu}, \Sigma)$ is the prior.
The variational distribution is set to $q_\phi(\boldsymbol{\mu}, \Sigma \mid \rvx) = \mathcal{K}_c(\alpha_\phi(\rvx), \beta_\phi(\rvx), \gamma_\phi(\rvx))$, where $\alpha_\theta(\rvx) \in \R^n$ and $\beta_\phi(\rvx), \gamma_\phi(\rvx) \in \R^n_{>0}$, and the prior is set to $p(\boldsymbol{\mu}, \Sigma) = \mathcal{K}_c(\boldsymbol{0}, I, I)$ in our experiments given curvature $-c$.
The pseudo-algorithm for the decoder of GM-VAE is present at Algorithm~\ref{alg:dec}. 

\section{Related Work}

\paragraph{Information geometry on VAE.}

Focusing on the bridge between probability theory and differential geometry, the adaptation of information geometry to the deep learning framework has been investigated in various aspects~\citep{dnnGeo, seqGeo, oodGeo}.
Having said that, \citet{vaeGeo} show that the training process of VAE can be seen as minimizing the distance between the two statistical manifolds: manifolds with the parameters of the decoder and the encoder.
Not only can the parameters but the outputs from the VAE decoder be modeled as probability distributions. \citet{decGeo} suggest a method of using the pull-back metric defined with arbitrary decoders on the latent space.
Our work focuses more on the statistical manifolds lying on the outputs of the encoder with the benefits from the information geometry.

\paragraph{VAE with Riemannian manifold latent space.}

The latent space of VAE reflects the geometrical property of the representations of the data.
The efficacy of setting the latent space to be hyperbolic space~\citep{mathieu19,nagano19,rown} or elliptic space~\citep{sphericalVAE1,sphericalVAE2} has been verified for various datasets.
\citet{MVAE} further extend the approach to enable the latent space to be the product of Riemannian manifolds with different learnable curvatures.
On top of these, we explore the method of setting the latent space to be the diagonal Gaussian manifold, which can be viewed as the product of hyperbolic spaces, and provide a novel viewpoint on prior work with information geometry.

\paragraph{Distributions over hyperbolic space.}
Defining a tractable distribution over hyperbolic space is challenging.
\citet{nagano19} suggest hyperbolic wrapped normal distribution (HWN) from the observation that the tangent space is Euclidean space. 
Leveraging operations defined on the tangent spaces, e.g., parallel transport, enables an easy sampling algorithm.
\citet{mathieu19} propose a rejection sampling method for the Riemannian normal distribution defined on the Poincar\'e disk model, namely Poincar\'e normal distribution.
This method rejects the pathological samples and enables accurate sampling from the distribution in exchange for high computational complexity.

Although these distributions are widely adopted in many applications~\citep{MVAE,rcnf,rown}, one can barely adopt the full covariance matrix due to the difficulties in Monte-Carlo based KL approximation.
The number of samples to approximate the KL divergence increases exponentially when the full covariance matrix is used~\citep{rown}, so it is common to use isotropic or diagonal covariance instead.
Especially in the Poincar\'e normal, the computation of KL divergence is slow due to the expensive rejection sampling.

% These distributions are studied in many cases~\citep{MVAE,rcnf,rown} but suffer from instability because of the absence of closed-form KL divergence.
% Our proposed distribution, however, not only shares the common merits but also overcomes the stability problem with closed-form KL divergence.
% Our method enjoys easy sampling and provides closed-form KL divergence while utilizing the geometric structure of the statistical manifold, i.e., the use of (approximated) geodesic distance.

% \paragraph{Neural network architectures on the hyperbolic space.}
% Considering the points in Riemannian manifolds, as the operations are highly dependent on the type of manifold, adopting the normal weighted sum defined on Euclidean space may not be appropriate.
% In this sense, several works propose new neural network architectures with appropriate operations for the hyperbolic space, such as fully-connected layer~\cite{hnn,hnnplus,fhnn}, graph convolutional layer~\citep{hgcn}, and attention mechanism~\citep{hat}.
% Our work suggests a new architecture with operations defined on \emph{the product of hyperbolic spaces}, which is an extension of the single hyperbolic space, that enables the aggregation of the $n-$dimensional data points.

\paragraph{RL with hyperbolic space.}
The hierarchical relationship between the states lying on the trajectories earned from RL agents has been gaining attention recently.
\citet{nagano19} have studied that the hierarchical structure of Atari2600 Breakout game states can be well-captured with hyperbolic VAEs.
We compare the same task, where GM-VAE outperforms the previous work.
\citet{hyp_deep_rl} suggest using hyperbolic space as the geometric prior for representation learning in model-free RL agent, showing improvements in generalization performances.
Here, we focus on model-based RL, especially the method of using the world model~\citep{worldModel,DreamerV2}, and open the possibility of applying hyperbolic space to broader domains of RL by solving the bottleneck of the numerical stability.

\section{Experiments}
In this section, we demonstrate the performances of GM-VAE on two tasks: density estimation of image datasets and model-based RL. 
We remark on the practical properties of GM-VAE shown in the experiments with additional analyses.
% We apply GM-VAE to very deep VAE (VDVAE)~\cite{vdvae} for the image and to the DreamerV2 architecture~\cite{Dreamerv2} for the model-based RL task.

% compare the performance of GM-VAE with the three baselines: Euclidean VAE, hyperbolic wrapped normal VAE (HWN VAE), and Poincar\'e VAE.
% The Euclidean VAE is the standard VAE with Euclidean latent space. 
% The HWN VAE uses the product of two-dimensional Lorentz models as a latent space and uses the hyperbolic wrapped normal to model the prior and variational distributions.
% The Poincar\'e VAE uses the product of two-dimensional Poincar\'e disk models as a latent space and uses the Poincar\'e normal to model the prior and variational distributions.
% The Euclidean VAE, HWN VAE, and Poincar\'e VAE are denoted as $\mathcal{E}$-VAE, $\mathcal{L}$-VAE, $\mathcal{P}$-VAE, respectively in the following results. 

\begin{table*}[t!]
    \centering
    \caption{Density estimation on real-world datasets. $d$ denotes the latent dimension. We report the negative test log-likelihoods of average 10 runs for Breakout, CUB, Food101, and Oxford102 with 95\% confidence interval. N/A in the log-likelihood indicates that the results are not available due to the failure of all runs, and N/A in the standard deviation indicates the results are not available due to failures of some runs. The best results are bolded.}
    % \vskip 0.1in
    \resizebox{\textwidth}{!}{%
    \begin{tabular}{l l r r r r r r}
        \toprule
         &\multicolumn{1}{c}{$d$}&\multicolumn{1}{c}{$\mathcal{E}$-VAE}&\multicolumn{1}{c}{$\mathcal{L}$-VAE}&\multicolumn{1}{c}{$\mathcal{P}$-VAE} & \begin{tabular}{@{}c@{}}GM-VAE\\($c=1$)\end{tabular}&\begin{tabular}{@{}c@{}}GM-VAE \\ ($c=1/2$)\end{tabular}&\begin{tabular}{@{}c@{}}GM-VAE \\ ($c=3/2$)\end{tabular}\\
         \midrule
         \multirow{3}{*}{Breakout} 
         &2&$124.74_{\pm0.86}$&$122.58_{\textrm{N/A}}$&$270.05_{\pm2.84}$&$\pmb{121.52_{\pm1.00}}$&$122.64_{\pm1.13}$&$122.47_{\pm1.98}$\\
         &4&$66.39_{\pm0.76}$&$66.70_{\pm0.32}$&$271.73_{\pm42.95}$&$65.83_{\pm0.49}$&$66.39_{\pm0.50}$&$\pmb{65.80_{\pm0.49}}$\\
         &8&$\pmb{44.97_{\pm0.37}}$&$45.25_{\pm0.27}$&$81.55_{\pm64.61}$&$45.14_{\pm0.30}$&$45.31_{\pm0.36}$&$45.36_{\pm0.49}$\\
         \midrule
         \multirow{3}{*}{CUB} &50&$992.05_{\pm1.38}$&$993.03_{\pm1.64}$&$990.49_{\pm2.26}$&$985.46_{\pm3.82}$&$986.27_{\pm3.81}$&$\pmb{979.14_{\pm3.70}}$\\
         &60&$969.99_{\pm3.13}$&$968.79_{\pm3.70}$&$964.02_{\pm3.55}$&$958.00_{\pm3.25}$&$960.88_{\pm3.46}$&$\pmb{956.77_{\pm2.53}}$\\
         &70&$949.13_{\pm2.72}$&$948.88_{\pm3.19}$&$944.24_{\pm4.40}$&$939.08_{\pm3.12}$&$942.34_{\pm3.44}$&$\pmb{937.15_{\pm2.76}}$\\
         \midrule
         \multirow{3}{*}{Food101} 
         & 50& $1297.81_{\pm4.51}$& $1298.45_{\pm6.32}$& $1293.26_{\pm7.14}$& $\pmb{1286.30_{\pm6.19}}$& $1299.58_{\pm7.02}$& $1290.57_{\pm8.23}$\\
& 60& $1224.03_{\pm8.31}$& $1227.16_{\pm5.18}$& $1218.09_{\pm3.88}$& $1213.31_{\pm3.88}$& $1216.63_{\pm4.56}$& $\pmb{1207.30_{\pm5.12}}$\\
& 70& $1164.95_{\pm3.80}$& $1165.39_{\pm5.54}$& $1165.91_{\pm4.91}$& $1152.80_{\pm3.35}$& $1160.97_{\pm4.18}$& $\pmb{1149.56_{\pm3.41}}$\\
         \midrule
         \multirow{3}{*}{Oxford102}
         &50&$1297.41_{\pm2.69}$&$1296.41_{\pm1.56}$&$1294.12_{\pm1.80}$&$1292.90_{\pm3.43}$&$\pmb{1289.43_{\pm2.46}}$&$1289.99_{\pm1.72}$\\
         &60&$1253.80_{\pm2.57}$&$1256.52_{\pm2.99}$&$1251.77_{\pm1.82}$&$\pmb{1245.49_{\pm2.18}}$&$1248.72_{\pm1.62}$&$1247.47_{\pm2.51}$\\
         &70&$1231.52_{\pm3.18}$&$1229.38_{\pm3.44}$&$1219.75_{\pm1.72}$&$1215.07_{\pm2.52}$&$1218.54_{\pm3.85}$&$\pmb{1214.85_{\pm2.56}}$\\
         \bottomrule
    \end{tabular}%
    }
    \label{tab:density_estimation_results}
\end{table*}

\subsection{Density estimation on image datasets}
We conduct density estimation on image datasets to measure the effectiveness of hyperbolic latent space against Euclidean space with the proposed GM-VAE.
% Dataset specifications.
We use three datasets: the images from Atari2600 Breakout with binarization (Breakout)~\citep{nagano19}, Oxford 102 Flower (Oxford102)~\citep{oxford102}, Food101~\citep{food101}, and Caltech-UCSD Birds-200-2011 (CUB)~\citep{CUB}.
The datasets are chosen with the four lowest $\delta$-hyperbolicity ($\delta$-H), a metric that measures how the given images are well-embed in hyperbolic space. Low $\delta$-H implies that the dataset is likely to embed in hyperbolic space.
The details about $\delta$-H are available in \autoref{apx:dh}.
The values of $\delta$-H for the four datasets and other candidate datasets are in Appendix~\ref{apx:dh_image}.
Several studies show that the images from the chosen datasets have an implicit hierarchical structure~\citep{nagano19, bird_hierarchy, food101, flower_hierarchy}.

% We display sample images from CUB and Oxford102 in \autoref{fig:image_hierarchy} with the implicit hierarchical structure studied in several studies~ \citep{bird_hierarchy,flower_hierarchy}.

% binarized-MNIST~\citep{mnist}, binarized-Omniglot~\citep{omniglot}, and the images from Atari 2600 Breakout with binarization (binarized-Breakout)~\citep{nagano19}.
% The binarized-Breakout is collected from plays with a pre-trained Deep Q-Network~\citep{deepq}.
% The size of images of all the datasets is resized to $64 \times 64$, while the images of Atari 2600 Breakout are binarized with a threshold value of 0.1; the threshold for Breakout is determined to visualize the components clearly.
% The size of images are $28 \times 28$, $28 \times 28$, and $80 \times 80$ for binarized-MNIST, binarized-Omniglot, and binarized-Breakout, respectively.
% The value of the threshold for binarization is set to 0.5, 0.5, and 0.1 for binarized-MNIST, binarized-Omniglot, and binarized-Breakout, respectively; the threshold for binarized-Breakout is determined to visualize the components clearly.

% Baseline

We compare GM-VAE with the three baseline models: VAE with Euclidean latent space ($\mathcal{E}$-VAE), and hyperbolic VAE equipped with HWN ($\mathcal{L}$-VAE) and Poincar\'e normal ($\mathcal{P}$-VAE). 
We use the product latent space for both $\mathcal{L}$-VAE and $\mathcal{P}$-VAE, and set the curvature value to $-1$.
The other details on the implementation and experimental setups are described in Appendix~\ref{apx:density_estimation_details}.

% Results
The results are reported at \autoref{tab:density_estimation_results}.
GM-VAE outperforms the baselines in all the settings, except one case of Breakout.
Especially in CUB and Oxford102, GM-VAE outperforms the baselines regardless of the curvature value.
In Breakout, $\mathcal{P}$-VAE shows inferior performance due to unstable training, and $\mathcal{L}$-VAE fails in some of the runs with small latent dimension.
The results of $\mathcal{P}$-VAE and $\mathcal{L}$-VAE with non-product latent space, a common choice in previous work, are also present in \autoref{apx:non_product}.

\vspace*{-0.1cm}

% \input{tables/geometric_transformations}

% \paragraph{Geometric transformations}
% We conduct an ablation study on the geometric transformations of GM-VAE.
% We compare the setting of GM-VAEs incorporating the geometric transformations to the setting of GM-VAEs using only an exponential function to send the output of the encoder to the Gaussian manifold but no additional geometric transformation at the first layer of the decoder.
% The results are in \autoref{tab:architecture_ablation}.
% We can see that the geometric transformations enhance the performance of the GM-VAE, except for two results that have similar performance.

\subsection{State representation learning in model-based RL}
We focus on the model-based RL task to verify the utility of GM-VAE on various tasks. 
Specifically, we apply GM-VAE to a world model, which aims to learn the representation of the environments~\citep{worldModel,Dreamer,Planet,DreamerV2}.
We use DreamerV2~\citep{DreamerV2} as the baseline model to evaluate the performance of GM-VAE in modeling environments.
DreamerV2 is composed of a recurrent state space model (RSSM)~\citep{Planet} and three predictors for the image $p_\phi(x_t|h_t, z_t)$, the reward $p_\phi(r_t|h_t, z_t)$, and the discount factor $p_\phi(\gamma_t|h_t, z_t)$, where $x_t$ is the observation which the format is the image, $r_t$ is the reward, $\gamma_t$ is the discounting factor, $h_t$ is the deterministic recurrent state, and $z_t$ is the stochastic state.
The model is trained by maximizing the likelihood of $p(\rvx, \rvr, \pmb{\gamma} \mid \rva)$ given observations $\rvx$, rewards $\rvr$, and discount factors $\pmb{\gamma}$ earned from the sequence of actions $\rva$ of an agent.
By deriving the evidence lower bound of $p(\rvx, \rvr, \pmb{\gamma} \mid \rva)$, the world model is learned to optimize the likelihood with the variational distribution $q_\theta(z_t \mid h_t, x_t)$ with the following objective $\mathcal{L}(\phi, \theta)$ as:
\begin{align*}
\mathcal{L}(\phi, \theta) = 
    \mathbb{E}%_{q_\theta}
        \Bigg[ 
        \sum_{t=1}^T \Big(
            -\log p_\phi(x_t, r_t, \gamma_t | h_t, z_t)
            %- \log p_\phi (r_t | h_t, z_t)
            %- \log p_\phi (\gamma_t | h_t, z_t) 
            + \beta \operatorname{D}_{\text{KL}}\left[ q_\theta(z_t | h_t, x_t) \parallel p_\phi(z_t | h_t) \right]
        \Big)
        \Bigg],
\end{align*}
where $\beta$ is KL loss scaling factor, $T$ is the length of input sequence, $p_\phi$ is the prior, and $q_\phi$ is the approximated posterior. 
GM-VAE is employed by replacing the space of $z_t$ with the Gaussian manifold and two components in RSSM, the representation model $q_\theta(z_t | h_t, x_t)$ and transition predictor $p_\phi(z_t | h_t)$, with PGM normal.

% Baseline
% The progress \cite{DreamerV2} have made in the field of world model learning is to replace the Euclidean latent space with the discrete latent space.
% This enhances the performances on many of the games in the Atari 2600 environment~\cite{atari} compared to the Euclidean baseline.
%, while the theoretical background on such alteration has not been understood yet.

% \input{figures/image_hierarchy.tex}

% Setting
We compare evaluation scores between different types of latent space on world model learning over the Atari2600 environments.
The agents are trained with 100M environment steps. 
We select games having the $\delta$-H values of the four lowest and the two highest among 60 popular Atari2600 games.
%The $\delta$-H for all 60 games are available in Appendix~\ref{apx:dh_atari}.
The other details on the implementation and experimental setups are described in Appendix~\ref{apx:rl_details} with the $\delta$-H for all 60 games in Appendix~\ref{apx:dh_atari}.

\begin{figure}[t!]
    \centering
    % \vskip -0.1in
    \begin{subfigure}[b]{.285\textwidth}
        \includegraphics[width=\textwidth, clip, trim=0cm 0.7cm 0cm 0cm]{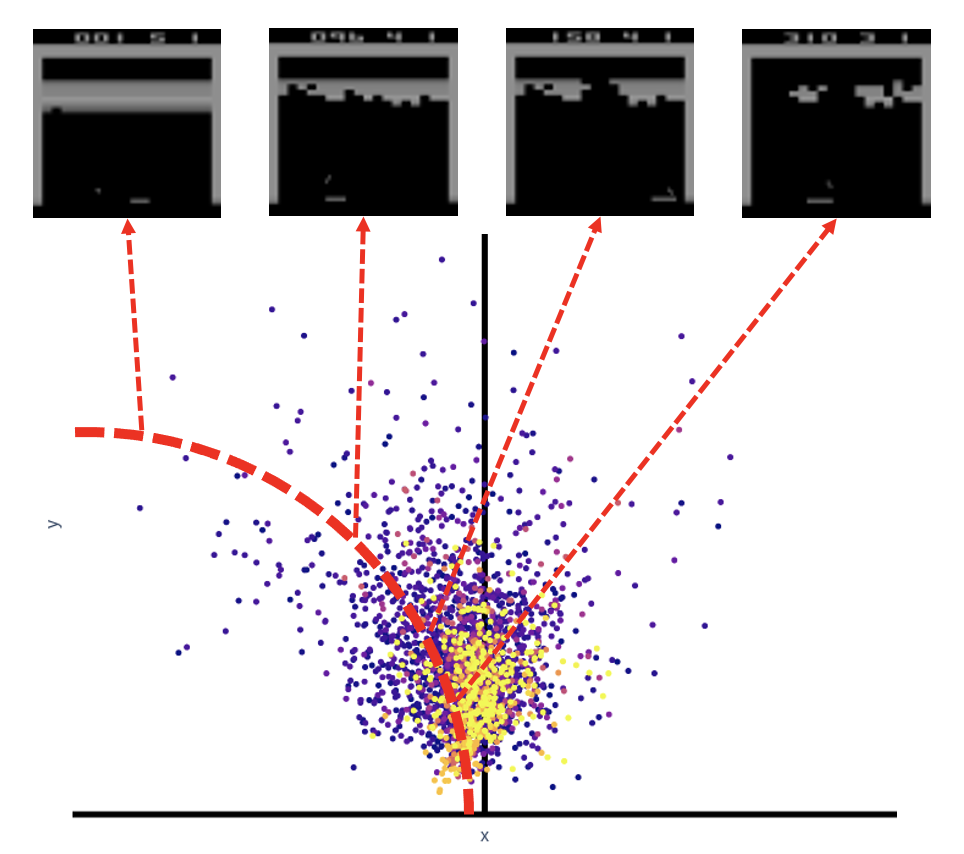}
        % \label{fig:world_model_analysis}
        \caption{\label{fig:world_model_analysis}Latent space analysis}
    \end{subfigure}
    \hspace{3mm}
    \begin{subfigure}[b]{.6\textwidth}
    \centering
    % \adjustbox{valign=b}{
    % \resizebox{\textwidth}{!}{
    
        \begin{tabular}{c r r r r}
        \toprule
         Latent space & Euc. & Disc. & Hyp. & $\delta-$H \\
         \midrule
         Breakout & $\mathbf{329.0}$ & 256.8 & 319.3 & 0.12 \\
         Alien & 3412.5 & 3120.0 & $\mathbf{3485.0}$ & 0.14 \\  
         Zaxxon & 34275 & 38825 & $\mathbf{38950}$ & 0.14 \\
         Ice Hockey & $\mathbf{25.50}$ & 11.80 & 20.75 & 0.14 \\
         \midrule
         Freeway & 32.8 & $\mathbf{33.0}$ & $\mathbf{33.0}$ & 0.38 \\
         Krull & 53290 & 36135 & $\mathbf{66185}$ &  0.38 \\
         \bottomrule
    \end{tabular}
    %\vspace{3.3mm}
    % }}
    
    \caption{\label{tab:rl_results}Results}
    \end{subfigure}
    
    \caption{
    The results of model-based RL experiment.
    (a) The dots from yellow to purple represent the latent states from the world model in the Atari2600 Breakout with decreasing rewards. 
    Along the red geodesic dashed line passing, we sample for images to visualize the learned representations. 
    As the sample shows, we can observe a hierarchical structure at different stages of the game along the geodesic.
    (b) We compare the methods of using Euclidean, discrete, and hyperbolic latent space.
    We report averaged rewards over four runs and bold the best reward.
    }
    % \vspace*{0.7cm}
    % \label{fig:world_model_analysis}
\end{figure}

% Results
With a commonly-used hyperbolic distribution, i.e., HWN, we observe that training the world model fails due to the numerical stability issue.
On the other hand, GM-VAE shows competitive results with the baselines in Euclidean and discrete latent space in all the games we test.
The results are reported in Figure \ref{tab:rl_results}.
We note that the reproduced Euclidean baseline results by using the official code are better than those reported in \citet{DreamerV2}.

\subsection{Remark on GM-VAE}

\paragraph{Numerical stability.}
One notable property of GM-VAE is the numerical stability during training compared to $\mathcal{L}$-VAE and $\mathcal{P}$-VAE. 
During the experiments, $\mathcal{L}$-VAE and $\mathcal{P}$-VAE fail to run in some of the Breakout image density estimations and all the seeds of model-based RL due to the numerical instability. Similar observations are also reported in several previous works~\citep{mathieu19, fhnn, MVAE}.
The sampling from a hyperbolic distribution is a major cause of the numerical instability. Consequently, the sampling-based KL divergence computation can be unstable.

% We first show that sampling from PGM normal can be stably done by a simple reparameterization.
% Sampling $\mu, \sigma$ from PGM normal $\mathcal{K}_c(\alpha, \beta, \gamma)$ is done by sampling $\mu$ from $\mathcal{N}(\alpha, \beta^2\gamma^2)$ and $\sigma^2$ from $\textrm{Gamma}(a, b)$ where $a = 1/4c\gamma^2 + 1, b=1/4c\beta^2\gamma^2$ as explained at Appendix \ref{apx:sampling_details}.
% We can sample from $\textrm{Gamma}(a, b)$ by using a reparameterization trick: $\sigma^2 = \epsilon \cdot b$, where $\epsilon$ is sampled from $\textrm{Gamma}(a, 1)$.
% The numerical stability issue in sampling from hyperbolic distribution occurs when the parameter is out of the manifold due to the machine precision error.
% In the case of GM-VAE, when the output of the VAE encoder $\beta$ is out of the manifold, e.g., $\beta = 0$, numerical stability issue can occur in the computation of the rate parameter $b$.
% We replace the instable sampling with sampling $\log \sigma^2$ by applying logarithm to the reparameterization trick $\log \sigma^2 = \log \epsilon + \log b$ and modifying the VAE encoder output from $\beta, \gamma \in \R_{>0}$ to $\log \beta^2, \log \gamma^2 \in \R$.
% The modifications successfully remove the instable operations from the sampling procedure.

We first show that sampling from PGM normal can be stabilized via a simple reparameterization trick.
To train GM-VAE, one needs to obtain sample $\mu$ and $\sigma$ from PGN normal $\mathcal{K}_c(\alpha, \beta, \gamma)$. Sampling $\mu$ can be done from $\mathcal{N}(\alpha, \beta^2\gamma^2)$ without numerical issues. Sampling $\sigma$ can be done from $\textrm{Gamma}(a, b)$ where $a = 1/4c\gamma^2 + 1, b=1/4c\beta^2\gamma^2$ as shown in Appendix \ref{apx:sampling_details}. However, due to machine precision error, often, $\beta$ violates the manifold constraints, i.e., $\beta = 0$. Eventually, direct sampling of $\sigma$ can cause the numerical instability. 
To avoid $\beta$ being zero, we use the output of the VAE encoder as $\log \beta^2$ whose value ranges over the entire real numbers and is more stable even when $\beta$ is close to zero. With $\log \beta^2$, instead of sampling $\sigma^2$, we sample $\log \sigma^2 = \log \epsilon + \log b$, where $\epsilon$ is sampled from $\textrm{Gamma}(a, 1)$, through the reparameterization of the Gamma distribution, where $\log b$ can be directly computed from $\log \beta^2$. 

We can show that the KL divergence between an arbitrary PGM normal and prior distribution $\mathcal{K}_c(\boldsymbol{0}, I, I)$ has a closed-form solution without any sampling.
The KL divergence of PGM normal is the sum of the KL divergences between two Gaussian distributions and between two Gamma distributions, as shown in Appendix \ref{apx:kl_details}.
First, the KL divergence between a univariate Gaussian distribution $\mathcal{N}(\mu, \sigma^2)$ and the prior distribution can be obtained with $\log \sigma^2$ as shown in \autoref{eq:kl_divergence}.
Second, the KL divergence between the two Gamma distributions, $\textrm{Gamma}(a_1, b_1)$ and $\textrm{Gamma}(a_2, b_2)$, written as:
\begin{equation*}
    \KL(\textrm{Gamma}(a_1, b_1) \parallel \textrm{Gamma}(a_2, b_2)) = a_2 \log \frac{b_1}{b_2} - \ln \frac{\Gamma(a_1)}{\Gamma(a_2)} + (a_1 - a_2) \psi(a_1) - \left(1 - \frac{b_2}{b_1}\right) a_1,
\end{equation*}
where $\psi$ is the digamma function, can be computed using $\log b$.

% reason: L-VAE
% The HWN uses the exponential map and parallel transport to transform the output of the encoder to the Lorentz model during the sampling.
% The sampling process of $\rvz$ from HWN follows:
% \begin{equation}
%     \rvv \sim \mathcal{N}(\boldsymbol{0}, \Sigma) 
%     \quad
%     \rvz = \exp_{\boldsymbol{\mu}}(\textrm{PT}_{\boldsymbol{0}_{\mathcal{L}} \rightarrow \boldsymbol{\mu}}([0, \rvv]))
% \label{eq:wrapped_normal}
% \end{equation}
% where $\boldsymbol{0}_{\mathcal{L}}$ is the origin of the Lorentz model $\mathcal{L}$, $\boldsymbol{\mu} \in \mathcal{L}$ is the mean vector of the distribution, and $\rvv$ is a sample obtained from the Gaussian distribution with zero mean and covariance $\Sigma$.
% The numerical instability of HWN often comes from the exponential map used in the Lorentz model.

% $\mathcal{P}$-VAE suffers from unstable training originating from the absence of closed-form KL divergence and the numerical instability of the Poincar\'e disk model~\citep{mathieu19,MVAE}.
% To train $\mathcal{P}$-VAE, the KL divergence between the Poincare normal is approximated based on the log probability of the samples. 
% The computation of the log probability, here, often becomes unstable when $\rvx, \rvy$ becomes closer to the boundary of the open disk, as the denominator of the Poincar\'e geodesic distance between $\rvx, \rvy$ in converges to zero.

% The PGM normal, on the other hand, does not suffer from instability with the help of stability when using log-covariance. 
% Please check the detailed arguments in \autoref{apx:numerical_stability}.

\paragraph{Training time comparison.}

\begin{wraptable}{r}{.5\textwidth}
    \centering
    % \vskip -0.19in
    \caption{
    The training time of the VAEs in density estimation of the Breakout image dataset. 
    We report the training time of the VAEs in seconds per epoch.
    GM-VAE is 1.93x faster than $\mathcal{P}$-VAE and 1.41x faster than $\mathcal{L}$-VAE in the experiments held on a single A100 40GB PCI GPU.
    }
    \label{tab:training_time}
    % \vskip 0.1in
    % \resizebox{\linewidth}{!}{
    \begin{tabular}{c c c c}
        \toprule
        $\mathcal{E}$-VAE & $\mathcal{L}$-VAE & $\mathcal{P}$-VAE & GM-VAE \\
        \midrule
        24.5 & 35.9 & 49.2 & 25.5 \\
        \bottomrule
    \end{tabular}
    % }
    % \vspace{-0.2in}
\end{wraptable}

Common bottlenecks of the mode hyperbolic VAEs arise from the complex manifold constraints and the difficulty of sampling from the hyperbolic distributions.
For example, the Poincar\'e disk model of $\mathcal{P}$-VAE and the Lorentz model of $\mathcal{L}$-VAE requires the samples to be inside of a unit disk and to be on a hyperboloid with constraint $\{\rvx \in \R^{n+1} \mid -x_0^2+\sum_{i=1}^n x_i^2\}$, respectively.
Such manifolds need complex transformations, e.g., clipping, projection, or geometric transformations using the Riemannian operations, to match the manifold constraint so making the training of the hyperbolic VAEs slower.
The Gaussian manifold, on the other hand, has a much simple manifold constraint and even does not require any transformations if we utilize the log space of $\sigma$.

% Common bottlenecks of the previous hyperbolic VAEs arise from geometric transformations to match various manifold constraints and from the difficulty of sampling from hyperbolic distributions. 
% at the last layer of the encoder, the first layer of the decoder, and sampling from the distribution.
%Both $\mathcal{P}$-VAE and $\mathcal{L}$-VAE require additional transformations at the output of the encoder and the input of the decoder to match the manifold constraint.
% For example, HWN of $\mathcal{L}$-VAE uses geometric transformations in the sampling procedure, and Poincar\'e normal of $\mathcal{P}$-VAE uses rejection sampling, which demands additional complexity in training time.
% GM-VAE, however, does not need additional transformations in the encoder, the decoder, and the sampling procedure. 
% The only manifold constraint of the Gaussian manifold is that $\beta$ and $\gamma$ of the PGM normal have to be positive, which can be easily achieved using exponential or softplus function.
% Furthermore, thanks to the factorization \autoref{eq:factorization}, sampling from the PGM normal is held directly on well-known distributions such as Gaussian and Gamma distributions, which is same as the Euclidean VAE.

We report the time consumptions of the VAEs with the latent dimension of 8 per epoch in the density estimation of Breakout at \autoref{tab:training_time}.
The results demonstrate that the algorithmic distinctions enable GM-VAE to be trained much faster than the baseline hyperbolic VAEs and even similar to $\mathcal{E}$-VAE.

\paragraph{Latent space analysis.}
% Breakout in Model-based RL
% We report the qualitative analysis of the learned representation in the hyperbolic space and probe the difference from the representation with Euclidean space in the model-based RL task \dw{(there is no comparison provided here. are we going to add one?)}.

% Qualitative result
We present a plot of the learned representation in the hyperbolic space at \autoref{fig:world_model_analysis} for qualitative analysis.
We take the world model with GM-VAE trained for Breakout and illustrate the geodesic starting from the origin in the figure with four generated samples along with the geodesic. 
We also provide the scatter plot of game states with their cumulative rewards represented in different colors. 
The brighter the color, the higher the cumulative reward.
The scatter plot reveals that the states with high cumulative rewards are distributed near the origin. 
Together with the samples from the geodesic, we can observe that the hierarchical structure of Breakout is well captured in the latent space.

Note that in the Poincar\'e disk model, the depth of the hierarchy is expected to be shown as the distance from the origin~\citep{poincare_embedding}.
When the root node is placed near the origin, the leaf nodes are likely to be placed near the boundary of the open disk.
The geodesic lines starting from the origin to the boundary of the Poincar\'e disk model are identical to the geodesics of the Gaussian manifold starting from $(0, 1)$, i.e., the origin of the Gaussian manifold.
The connection implies that the data hierarchy should be aligned along the geodesic curves if the hierarchy is well captured. %starting from the origin and having a shape similar to fountain water but a piece of an ellipse.

% Quantitative result
To quantitatively measure the correlation between the cumulative rewards and the states, we measure the Pearson correlation between the cumulative reward and the norm of the states. We obtain a correlation coefficient of 0.46 from the hyperbolic latent space, whereas the correlation coefficient of the Euclidean latent space is 0.40, showing the hyperbolic space better captures the hierarchy along the increasing norm. More experimental details are explained in Appendix~\ref{apx:rl_analysis}.

% \jylee{
% Measuring the correlation, between the norm of the representations from the world model with hyperbolic latent space and the cumulative rewards in Breakout, results in a value of 0.46.
% The correlation value, on the other hand, is 0.40 computed with the Euclidean latent space. More details on the procedure of measuring the correlation are explained in Appendix~\ref{apx:implementation_detail}.
% }

%\input{07_FutureWork.tex}
\section{Conclusion \& Future Work}

In this work, we propose a novel method of representation learning with GM-VAE, utilizing the Gaussian manifold for the latent space.
With the newly-proposed PGM normal defined over the Gaussian manifold, which shows better stability and ease of sampling compared to the commonly-used ones, we verify the efficacy of our method on several tasks.
Our method achieves outperforming results on density estimation with image datasets and competitive results on model-based RL compared to the baselines.
We explain the behavior of GM-VAE in terms of solving the frequent numerical issue of commonly-used hyperbolic VAEs.
The analysis of latent space exhibits that the hierarchy lying in the dataset can be preserved by using GM-VAE.

We suggest that the numerical stability of our method can be helpful for scaling the generative models, e.g., very deep VAE~\citep{vdvae}, endowed with hyperbolic geometrical priors.
As GM-VAE is beneficial for capturing hierarchy with promising results in modeling RL environment, another potential future work can be extending the use of hyperbolic space, such as learning a skill tree for solving complex long-horizon tasks~\citep{skimo}.
We believe that the connection between the statistical manifold and hyperbolic space provides new insight to the research community and hope to see more interesting connections and analyses in the future.

\begin{ack}
    This work was partly supported by Institute of Information \& communications Technology Planning \& Evaluation (IITP) grant funded by the Korea government (MSIT) (No.2019-0-01906, Artificial Intelligence Graduate School Program (POSTECH)) and National Research Foundation of Korea (NRF) grant funded by the Korea government (MSIT) (No. RS-2023-00217286) and National Research Foundation of Korea (NRF) grant funded by the Korea government (MSIT) (NRF-2021R1C1C1011375)
\end{ack}

\newpage
\bibliography{references}
\bibliographystyle{icml2023}

\newpage
\appendix
\onecolumn
\section{Machine Precision Error Analysis on Hyperbolic Space}

\begin{figure}[h!]
    \centering
      \includegraphics[width=.6\linewidth]{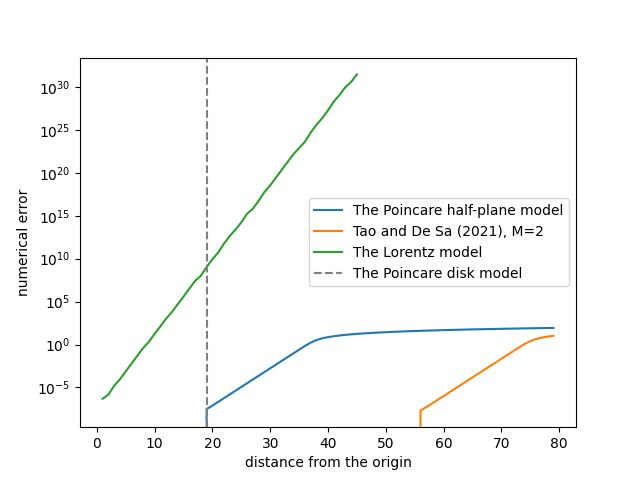}
      \caption{The machine precision errors of the hyperbolic models. For the Poincare half-plane model, we report the upper bound of the machine precision error derived by \citet{machineprecisionerror1}. For the Lorentz model, we report the error between the Lorentzian product of the point and -1, which is the manifold constraint of the Lorentz model. For the Poincare disk model, we report the threshold that the norm of the point becomes one. The precision is set to 32 bits. The analysis reveals that all the three models suffer from numerical stability issue with points which the distance between the origin is farther than 20.}
      \label{fig:machine_precision_error}
\end{figure}
\section{Gaussian Manifold}
\subsection{Curvature of the Gaussian manifold}
\label{apx:fisher_extension}
We construct a Riemannian manifold $\{(\mu, \sigma) \mid \mu \in \R, \sigma \in \R_{>0}\}$ with a positive constant $c$ and the metric tensor $\sigma^{-2}\textrm{diag}(1, 1/c)$, which we will name Gaussian manifold.
We need to show the value of the curvature.
% For the simplicity, we will denote the the manifold of the Gaussian manifold as $\{(\mu, t) \mid \mu \in \R, t \in \R_{>0}\}$ and the metric tensor of the Gaussian manifold as $\textrm{diag}(1/t, 1/(4ct))$.

First, we need to compute the Christoeffel symbols of the Gaussian manifold defined as:
\begin{equation*}
    \Gamma_{ij}^k = \frac{1}{2} g^{kl} \left( \frac{\partial g_{jl}}{\partial g_i} + \frac{\partial g_{il}}{\partial g_j} - \frac{\partial g_{ij}}{\partial g_l} \right),
\end{equation*}
where $g_{ij}$ is the $(i, j)$ element of the metric tensor and $g^{ij}$ is the $(i, j)$ element of the inverse of the metric tensor.

The Christoeffel symbols of the Gaussian manifold are:
\begin{align*}
    \Gamma_{ij}^1 &= \begin{pmatrix} 0 & -\frac{1}{\sigma} \\ -\frac{1}{\sigma} & 0 \end{pmatrix} \\
    \Gamma_{ij}^2 &= \begin{pmatrix} \frac{c}{\sigma} & 0 \\ 0 & -\frac{1}{\sigma} \end{pmatrix}.
\end{align*}

Then, the sectional curvature of the space $\kappa_g$ with given tangent vectors $du, dv$ is computed as:
\begin{align*}
    \kappa_g &= \frac{\langle R(d\mu, d\sigma)d\sigma, d\mu\rangle}{\det g} \\
    &= \frac{1}{\det g} \cdot g_{1m} \left( \frac{\partial \Gamma_{22}^m}{\partial \mu} - \frac{\partial \Gamma_{12}^m}{\partial \sigma} + \Gamma_{22}^p \Gamma_{1p}^m - \Gamma_{12}^p \Gamma_{2p}^m \right) \\
    &= \frac{-\frac{1}{\sigma^4}}{\frac{1}{c\sigma^4}} \\
    &= -c.
\end{align*}
Note that the sectional curvature of two-dimensional Riemannian manifold is same as the Gaussian curvature where $\langle d\mu, d\mu \rangle \langle d\sigma, d\sigma \rangle - \langle d\mu, d\sigma \rangle^2 = \det g$.
% where $Rm(u,v,v,u) := \langle R(u, v)v, u \rangle$, with given tangent vectors $du \;\&\; dv$, can be computed as:
% \begin{equation}
%     Rm(\mu, \sigma, \sigma, \mu) = g_{1m} \left( \frac{\partial \Gamma_{22}^m}{\partial \mu} - \frac{\partial \Gamma_{12}^m}{\partial \sigma} + \Gamma_{22}^p \Gamma_{1p}^m - \Gamma_{12}^p \Gamma_{2p}^m \right).
% \end{equation}
% By putting the metric tensor and the Christoeffel symbols together, the curvature of the Gaussian manifold is computed as:
% \begin{equation}
%     \kappa_g = \frac{Rm(\mu, \sigma, \sigma, \mu)}{\det g} = \frac{-\frac{1}{\sigma^4}}{\frac{1}{c\sigma^4}} = -c.
% \end{equation}

\subsection{Gaussian manifold with KL-divergence}
\label{apx:kl_gm}
Between two univariate Gaussian distributions $\mathcal{N}(\mu_1, \sigma^2_1)$ and $\mathcal{N}(\mu_2, \sigma^2_2)$, we can compute the KL divergence as:
\begin{equation}
\label{eq:kl_divergence}
    \KL(\mathcal{N}(\mu_1, \sigma^2_1) \parallel \mathcal{N}(\mu_2, \sigma^2_2)) = \frac{1}{2} \left(\log \frac{\sigma_2^2}{\sigma_1^2} + \frac{\sigma_1^2 + \left(\mu_1 - \mu_2\right)^2}{\sigma_2^2} - 1 \right).
\end{equation}

We extend the KL divergence for an arbitrary curvature of the Gaussian manifold as:
\begin{equation*}
    \KL^{\mathcal{G}_c}((\mu_1, \sigma_1), (\mu_2, \sigma_2)) := \frac{\KL(\mathcal{N}(\sqrt{2c}\mu_1, \sigma^2_1) \parallel \mathcal{N}(\sqrt{2c}\mu_2, \sigma^2_2))}{2c}.
\end{equation*}

Now, we show that the extended KL divergence still approximates the Riemannian distance of the manifold as:

\begin{align*}
    \KL^{\mathcal{G}_c}((\mu + d\mu, \sigma + d\sigma), (\mu, \sigma)) &= \frac{1}{2 \cdot 2c} \left(\log \frac{\sigma^2}{(\sigma + d\sigma)^2} + \frac{(\sigma + d\sigma)^2 + 2c(d\mu)^2}{\sigma^2} - 1\right) \\
    &= \frac{1}{2 \cdot 2c} \left(-2\log\left(1 + \frac{d\sigma}{\sigma}\right) + \frac{2\sigma d\sigma + (d\sigma)^2}{\sigma^2} + \frac{2c(d\mu)^2}{\sigma^2}\right) \\
    &= \frac{1}{2 \cdot 2c} \left(-2\left(\frac{d\sigma}{\sigma} - \frac{(d\sigma)^2}{2\sigma^2}\right) + \frac{2\sigma d\sigma + (d\sigma)^2}{\sigma^2} + \frac{2c(d\mu)^2}{\sigma^2} + \mathcal{O}((d\sigma)^3)\right)  \\
    &= \frac{1}{2}\begin{pmatrix}d\mu \\d\sigma \\\end{pmatrix}^T \begin{pmatrix}\frac{1}{\sigma} & 0 \\0 & \frac{1}{c\sigma^2} \\\end{pmatrix} \begin{pmatrix}d\mu \\d\sigma \\\end{pmatrix} + \mathcal{O}((d\sigma)^3).
\end{align*}

\section{Pseudo Gaussian Manifold Normal Distribution}
\label{apx:PGMNormal}

% \begin{align}
%     \frac{\KL(\mathcal{N}(\sqrt{2c}(\mu + d\mu), \sigma + d\sigma) \parallel \mathcal{N}(\sqrt{2c}\mu, \sigma))}{2c} &= \frac{(d\sigma)^2}{2c\sigma^2} + \frac{(d\mu)^2}{2\sigma^2} + \mathcal{O}((d\sigma)^3) \\
%     &= \frac{1}{2}\begin{pmatrix}d\mu \\d\sigma \\\end{pmatrix}^T \begin{pmatrix}\frac{1}{\sigma^2} & 0 \\0 & \frac{1}{c\sigma^2} \\\end{pmatrix} \begin{pmatrix}d\mu \\d\sigma \\\end{pmatrix} + \mathcal{O}((d\sigma)^3).
% \end{align}

% \subsection{Derivation of the probability density function}

In this section, we derive the normalizing constant $Z(c, \beta, \gamma)$ and the factorization of the PGM normal which the density function is defined as:
\begin{equation*}
    \mathcal{K}_c(\mu, \sigma; \alpha, \beta, \gamma) = \frac{\sigma^3}{Z(c, \beta, \gamma)}  \exp\left(-\frac{\KL^{\mathcal{G}_c}((\mu, \sigma), (\alpha, \beta))}{\gamma^2}\right).
\end{equation*}

\subsection{Normalizing Constant}
\label{apx:normalizing_constant_details}

The given probability density function needs to satisfy the following condition:
\begin{equation}
    \label{eq:pdf_constraint}
    \int_{\mathcal{G}_c} \mathcal{K}_c(\mu, \sigma; \alpha, \beta, \gamma) \sqrt{\det g} \cdot d(\mu, \sigma) = 1,
\end{equation}
where $\sqrt{\det g} \cdot d(\mu, \sigma)$ is the probability measure over the Gaussian manifold induced from the Lebesgue measure $d(\mu, \sigma)$.
The normalizing factor $Z(c, \beta, \gamma)$ can be computed using the condition \autoref{eq:pdf_constraint} as:
\begin{align}
    Z(c, \beta, \gamma)
    &= \int_0^\infty \int_{-\infty}^\infty \sigma^3 \cdot  \exp\left(-\frac{\KL^{\mathcal{G}_c}((\mu, \sigma), (\alpha, \beta))}{\gamma^2}\right) \frac{1}{\sqrt{c}\sigma^2} \, d \mu \, d\sigma \nonumber\\
    % &= \frac{1}{\sqrt{c}\beta^3}\int_0^\infty \int_{-\infty}^\infty \sigma \cdot \exp\left(-\frac{\KL^{\mathcal{G}_c}((\mu, \sigma), (\alpha, \beta))}{\gamma^2}\right) \, d \mu \, d\sigma\\
    % &= \int_0^\infty \int_{-\infty}^\infty  \exp\left(\frac{1}{4c\gamma^2}\left(\log \frac{\sigma^2}{\beta^2} - \frac{\sigma^2}{\beta^2} + 1\right) - \frac{(\mu - \alpha)^2}{2\beta^2\gamma^2}\right)\frac{1}{\sqrt{c}\sigma^2} \, d \mu \, d \sigma \\
    &= \frac{1}{\sqrt{c}}\left(\beta^{-\frac{1}{-2c\gamma^2}}\exp\left(\frac{1}{4c\gamma^2}\right)\int_0^\infty  \sigma \cdot (\sigma^2)^{\left(\frac{1}{4c\gamma^2} + 1\right) - 1}\exp\left(-\frac{\sigma^2}{4c\beta^2\gamma^2}\right)\, d\sigma\right) \nonumber\\
    & \quad\quad\quad\quad\quad\quad\quad\quad\quad\quad\quad\quad\quad\quad\quad\quad\quad\quad\quad\quad
    \times \left(\int_{-\infty}^\infty \, \exp\left(-\frac{(\mu - \alpha)^2}{2\beta^2\gamma^2}\right)d\mu\right) \nonumber\\
    % &= \left(\beta^{-\frac{1}{\gamma^2}}\exp\left(\frac{1}{2\gamma^2}\right) \Gamma\left(\frac{1}{2\gamma^2} + 1\right)(2\beta^2\gamma^2)^{\frac{1}{2\gamma^2}+1}\int_0^\infty \textrm{Gamma}(\frac{1}{2\gamma^2}+1, 2\beta^2\gamma^2)\, d\sigma^2\right) \left(\sqrt{2\pi}\beta\gamma \int_{-\infty}^\infty \mathcal{N}(\alpha, \beta^2\gamma^2)\, d\mu\right) \\
    &= \frac{1}{2\sqrt{c}}\sqrt{2\pi}\beta^3\gamma \exp\left(\frac{1}{4c\gamma^2}\right)\Gamma\left(\frac{1}{4c\gamma^2}\right)\left(\frac{1}{4c\gamma^2}\right)^{-\frac{1}{4c\gamma^2}} \nonumber \\
    & \quad\quad\quad\quad\quad\quad
    \times \left( \int_0^\infty \textrm{Gamma}\left(\sigma^2;\frac{1}{4c\gamma^2}+1, \frac{1}{4c\beta^2\gamma^2}\right)\, d\sigma^2 \right) \left( \int_{-\infty}^\infty \mathcal{N}(\mu; \alpha, \beta\gamma)\, d\mu\right) \label{eq:factorization_proof} \\
    % &= \beta^{-\frac{1}{\gamma^2}}\exp\left(\frac{1}{2\gamma^2}\right)\Gamma\left(\frac{1}{2\gamma^2} + 1\right)(2\beta^2\gamma^2)^{\frac{1}{2\gamma^2}+1}\sqrt{2\pi}\beta\gamma \\
    &=\frac{\sqrt{2\pi}\beta^3}{2\sqrt{c}} \gamma \exp\left(\frac{1}{4c\gamma^2}\right)\Gamma\left(\frac{1}{4c\gamma^2}\right)\left(\frac{1}{4c\gamma^2}\right)^{-\frac{1}{4c\gamma^2}}. \nonumber
\end{align}
Finally, the logarithm of the normalizing factor is computed as:
\begin{equation*}
    \log Z(c, \beta, \gamma) = \frac{1}{2} \log(2\pi) + 3\log \beta - \frac{1}{2} \log c - \log 2 + \frac{1}{2}\log \gamma^2 + \log \Gamma\left(\frac{1}{4c\gamma^2}\right) + \frac{1}{4c\gamma^2} (1 + \log (4c\gamma^2)).
\end{equation*}

\subsection{Sampling}
\label{apx:sampling_details}

Suppose that $p(\mu, \sigma) = \mathcal{K}_c(\mu, \sigma; \alpha, \beta, \gamma) \sqrt{\det g}$.
Sampling $\mu$ and $\sigma$ from the probability distribution $p(\mu, \sigma)$ can be done by sampling $\mu$ from the marginal distribution $p(\mu)$ and then sampling $\sigma$ from the conditional distribution $p(\sigma \vert \mu)$.
The marginal distribution $p(\mu)$ can be derived from \autoref{eq:factorization_proof} as:
\begin{align*}
    p(\mu) &= \int_{0}^{\infty} p(\mu, \sigma) \, d\sigma \\
    &= \int_{0}^{\infty} \mathcal{K}_c(\mu, \sigma; \alpha, \beta, \gamma) \sqrt{\det g} \, d\sigma \\
    &= \left( \int_0^\infty \textrm{Gamma}\left(\sigma^2;\frac{1}{4c\gamma^2}+1, \frac{1}{4c\beta^2\gamma^2}\right)\, d\sigma^2 \right)  \mathcal{N}(\mu; \alpha, \beta^2\gamma^2) \\
    &= \mathcal{N}(\mu; \alpha, \beta^2\gamma^2).
\end{align*}
\autoref{eq:factorization_proof} also implies that $\mu$ and $\sigma$ are independent in the aspect of $p(\mu, \sigma)$ so the conditional distribution $p(\sigma \vert \mu)$ is identical to the marginal distribution $p(\sigma)$.
The marginal distribution $p(\sigma)$ is computed as:
\begin{align*}
    p(\sigma) &= \int_{-\infty}^{\infty} p(\mu, \sigma) \, d\mu \\
    &= \int_{-\infty}^{\infty} \mathcal{K}_c(\mu, \sigma; \alpha, \beta, \gamma) \sqrt{\det g} \, d\mu \\
    &= \left( 2\sigma  \cdot \textrm{Gamma}\left(\sigma^2;\frac{1}{4c\gamma^2}+1, \frac{1}{4c\beta^2\gamma^2}\right)\right) \left( \int_{-\infty}^\infty \mathcal{N}(\mu; \alpha, \beta^2\gamma^2)\, d\mu\right) \\
    &= 2\sigma  \cdot \textrm{Gamma}\left(\sigma^2;\frac{1}{4c\gamma^2}+1, \frac{1}{4c\beta^2\gamma^2}\right).
\end{align*}

Here, sampling $\sigma$ from $p(\sigma)$ can be easily replaced by the  procedure of sampling $\sigma^2$ from $p(\sigma^2)$, which is identical to $p(\sigma) / (2\sigma) = \textrm{Gamma}\left(\sigma^2;\frac{1}{4c\gamma^2}+1, \frac{1}{4c\beta^2\gamma^2}\right)$, and then applying square root to the sample $\sigma^2$ to get $\sigma$.

\subsection{KL Divergence}
\label{apx:kl_details}

Suppose that $p(\mu, \sigma), q(\mu, \sigma)$ are two different PGM normal multplied with $\sqrt{\det g}$.
As shown in Appendix \ref{apx:sampling_details}, $\mu$ and $\sigma$ are independent so the KL divergence between $p, q$ is same as $\KL(p(\mu) \Vert q(\mu)) + \KL(p(\sigma) \Vert q(\sigma))$.
The first term is the KL divergence between two normal distribution.
The second term is same with $\KL(p(\sigma^2) \Vert q(\sigma^2))$ due to the change-of-variable formula, so it is the KL divergence between two gamma distribution.

\newpage
\section{$\delta$-Hyperbolicity} \label{apx:dh}
In this section, we explain about $\delta$-hyperbolicity ($\delta$-H) and show the $\delta$-H values of the candidate datasets of our experiments.

\subsection{$\delta$-hyperbolicity}

\begin{wrapfigure}{r}{.5\textwidth}
    \centering
    %\vspace*{0.25cm}
    \vskip -0.2in
    \begin{subfigure}[t]{.35\linewidth}
        \centering
        \begin{tikzpicture}
            \node[anchor=south west] at (0, 0) {\includegraphics[width=\linewidth]{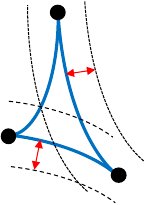}};
            \node[anchor=south west] at (0.375\textwidth, 1.4\textwidth) {\textsf{\tiny A}};
            \node[anchor=south west] at (0.025\textwidth, 0.55\textwidth) {\textsf{\tiny B}};
            \node[anchor=south west] at (0.775\textwidth, 0.3\textwidth) {\textsf{\tiny C}};
            \node[anchor=south west] at (0.7\textwidth, 0.95\textwidth) {\tiny \textcolor{red}{$\delta > 0$}};
        \end{tikzpicture}
        \caption{$\delta$-H $>0$}
    \end{subfigure}
    \hspace{5mm}
    \begin{subfigure}[t]{.35\linewidth}
        \centering
        \begin{tikzpicture}
            \node[anchor=south west] at (0, 0) {\includegraphics[width=\linewidth]{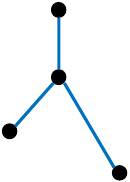}};
            \node[anchor=south west] at (0.425\textwidth, 1.4\textwidth) {\textsf{\tiny A}};
            \node[anchor=south west] at (0.05\textwidth, 0.475\textwidth) {\textsf{\tiny B}};
            \node[anchor=south west] at (0.9\textwidth, 0.15\textwidth) {\textsf{\tiny C}};
            \node[anchor=south west] at (0.525\textwidth, 0.8\textwidth) {\textsf{\tiny D}};
            \node[anchor=south west] at (0.7\textwidth, 0.95\textwidth) {\tiny \textcolor{red}{$\delta = 0$}};
        \end{tikzpicture}
        \caption{$\delta$-H $=0$}
    \end{subfigure}
    \caption{
    The illustration of $\delta$-H for given geodesic triangles.
    The blue lines denote the geodesic curves between the points.
    (a) When the side $AB$ is contained in the union of the two $\delta$-neighbor regions of the side $AC$ and $BC$, we say that the geodesic triangle is $\delta$-hyperbolic.
    $\delta$-H of the geodesic triangle is then defined as the minimum possible value of $\delta$.
    % Hyperbolic triangle satisfies: a side $AB$ is contained in the union of two $\delta-$neighborhood areas of the other two sides $AC$ and $BC$.
    (b) Any tree-structured triangle, where the geodesic curves correspond to the tree edges, is $0$-H. For example, given geodesic triangle $ABC$, one side $AB$ is already occupied by the two sides $AB$ and $BC$, as the geodesic between the points $A$ and $B$ is the union of edge $AD$ and $BD$ being occupied by other counterparts.%; $\delta$-H is equal to zero.
    }
    \label{fig:delta_hyperbolicity}
    \vskip -0.1in
\end{wrapfigure}

Given a metric space $(X, d)$, the metric space is said to be $\delta$-hyperbolic if, for any geodesic triangle, i.e., a triangle where each side is a geodesic curve, any point on the side of the geodesic triangle is within the distance of less than or equal to $\delta$ of the other two sides; when such $\delta$ exists, $(X, d)$ is said to be hyperbolic.
To be specific, when the Gromov product between any three points $x, y, z \in X$ is given as: 
\begin{equation*}
    (y, z)_x = \frac{1}{2}\left(d(x, y) + d(x, z) - d(y, z)\right),
\end{equation*}
if then the following inequality holds for any four points $x, y, z, w \in X$, we call the metric space is $\delta$-hyperbolic:
\begin{equation}
    (x, z)_w \geq \min((x, y)_w, (y, z)_w)) - \delta.
    \label{eq:dH}
\end{equation}
$\delta$-hyperbolicity ($\delta$-H) is defined to be the minimum value of $\delta$ satisfying \autoref{eq:dH} and is used as a measurement quantifying how a given metric space well embeds in hyperbolic~\cite{hyp_image_embedding,hyp_deep_rl}.
\autoref{fig:delta_hyperbolicity} illustrates the concept of $\delta$-H.
We note that the lower $\delta$-H the metric space has, the less deviation from the exact hyperbolic space is.

\paragraph{Computation.}
We measure the $\delta$-H values of the images $X$ from the datasets by following the procedure from \citet{delta_hyperbolicity_computation}.
We first extract the embeddings of the images using a pre-trained feature extractor to construct a metric space of the images.
We then randomly sample a fixed point $w$ and calculate the pairwise Gromov product of the embeddings $D$ with $w$ as \autoref{eq:dH}.
We finally determine the $\delta$-H of the images $X$ by finding the largest coefficient of $(\max_k \min_{ij} (D_{ik}, D_{kj})) - D$.

To reduce the scale difference between the datasets, we report the value $2\delta(X) / \textrm{diam}(X)$ where $\textrm{diam}(X)$ denotes the maximum pairwise distance of $X$.
Because computing the matrix $D$ among all the images $X$ is computationally expensive, we compute the $\delta$-H of randomly sampled 1,000 images from $X$.
We repeat this 10 times and report the average $\delta$-H.

\subsection{Image datasets}
\label{apx:dh_image}
We measure the $\delta$-H of the image datasets using an ImageNet pre-trained Inception V3 as a feature extractor.

\begin{table}[h!]
    \centering
    % \caption{TODO}
    % \label{tab:image_dh}
    % \vskip 0.1in
    % \resizebox{0.5\linewidth}{!}{
    \begin{tabular}{c c c c c c c}
        \toprule
        Breakout & CUB & Food101 & Oxford102  & CIFAR-10 & SVHN & CelebA 
        \\
        \midrule
        0.124 & 0.223 & 0.233 & 0.238 & 0.248 & 0.283 & 0.287 
        \\
        \bottomrule
    \end{tabular}
    %}
\end{table}

\subsection{Atari2600 environments}
\label{apx:dh_atari}
We collect the Atari2600 images using the pre-trained agents of \citet{atari_agents} and measure the $\delta$-H using the image encoder from the agents.
For each environment, we report the $\delta$-H of the images which are collected by the agent with the highest reward.
We also report the corresponding reward.
We run the agents for at least 6 episodes.

% Please add the following required packages to your document preamble:
% \usepackage{booktabs}
\begin{table}[t!]
\centering
\begin{tabular}{l | r r}
\toprule
Game             & $\delta$-H  & Reward       \\ \midrule
Breakout       & 0.12 & 340    \\
Alien          & 0.14 & 6855   \\
Zaxxon         & 0.14 & 13100  \\
IceHockey      & 0.14 & 3      \\
Gravitar       & 0.17 & 1004   \\
Carnival       & 0.17 & 5605   \\
RoadRunner     & 0.18 & 60050  \\
Pong           & 0.18 & 21     \\
Tutankham      & 0.18 & 236    \\
Boxing         & 0.19 & 99     \\
Solaris        & 0.19 & 1727   \\
WizardOfWor    & 0.20 & 17217  \\
Seaquest       & 0.20 & 29745  \\
Gopher         & 0.20 & 24173  \\
Centipede      & 0.20 & 4663   \\
PrivateEye     & 0.21 & 195    \\
Pitfall        & 0.21 & 0      \\
ElevatorAction & 0.21 & 64917  \\
Robotank       & 0.21 & 75     \\
MsPacman       & 0.21 & 5455   \\
DemonAttack    & 0.22 & 23876  \\
Asterix        & 0.22 & 26792  \\
Qbert          & 0.22 & 17364  \\
Jamesbond      & 0.22 & 886    \\
VideoPinball   & 0.23 & 632563 \\
UpNDown        & 0.23 & 27030  \\
FishingDerby   & 0.23 & 54     \\
Tennis         & 0.23 & 22     \\
Berzerk        & 0.24 & 904    \\
Riverraid      & 0.24 & 15103 \\
\bottomrule
\end{tabular}
% \hspace{0mm}
\quad
\begin{tabular}{l | r r}
\toprule
Game             & $\delta$-H  & Reward       \\ \midrule
AirRaid          & 0.25 & 11729  \\
Frostbite        & 0.25 & 8653   \\
Pooyan           & 0.25 & 7973   \\
ChopperCommand   & 0.26 & 10530  \\
KungFuMaster     & 0.26 & 27733  \\
YarsRevenge      & 0.26 & 56405  \\
Phoenix          & 0.27 & 30208  \\
JourneyEscape    & 0.27 & -707   \\
SpaceInvaders    & 0.27 & 15623  \\
Hero             & 0.28 & 28592  \\
Enduro           & 0.28 & 2089   \\
Assault          & 0.28 & 3217   \\
Venture          & 0.29 & 1529   \\
StarGunner       & 0.29 & 68417  \\
Atlantis         & 0.29 & 919750 \\
DoubleDunk       & 0.30 & 18     \\
TimePilot        & 0.30 & 10914  \\
BeamRider        & 0.30 & 7069   \\
BattleZone       & 0.30 & 40333  \\
NameThisGame     & 0.31 & 12925  \\
Skiing           & 0.31 & -10021 \\
Amidar           & 0.32 & 2826   \\
Asteroids        & 0.34 & 1405   \\
MontezumaRevenge & 0.34 & 0      \\
Bowling          & 0.35 & 49     \\
BankHeist        & 0.36 & 1563   \\
Kangaroo         & 0.37 & 14283  \\
CrazyClimber     & 0.37 & 132517 \\
Freeway          & 0.38 & 34     \\
Krull            & 0.38 & 9016   \\ \bottomrule
\end{tabular}
\end{table}
\newpage
\section{Implementation Details}
% \label{apx:implementation_detail}
In this section, we introduce the implementation details of the experiments.

\subsection{Density estimation on image datasets}
\label{apx:density_estimation_details}

% Dataset specs
% Image size, split size
We estimate the density of the images from Atari2600 Breakout~\citep{nagano19}, Oxford102~\citep{oxford102}, and CUB~\citep{CUB}.
The images of Breakout are collected from plays with a pre-trained Deep Q-Network~\citep{deepq}.
The size of images of all the datasets is resized to $64 \times 64$, while Breakout is binarized with a threshold value of 0.1; the threshold for Breakout is determined to visualize the components clearly.

We split the datasets into train, validation, and test.
For Breakout and CUB, we split the original train set into train and validation sets.
For Oxford102, because the original train set is too small, we merge the original train and test set and then split it into three splits.
For Food101, we randomly sample the train set and validation set from the original train set, and also randomly sample the test set from the original test set.
\begin{table}[h!]
    \centering
    \begin{tabular}{c | c c c c}
         \toprule
         Split & Breakout & CUB & Food101 & Oxford102 \\
         \midrule
         Train & 80,000 & 4,795 & 6,000 & 5,120 \\
         Validation & 9,503 & 1,199 & 1,000 & 1,228 \\
         Test & 9,934 & 5,794 & 1,000 & 1,025 \\
         \bottomrule
    \end{tabular}
\end{table}

% Architecture
% Encoder, decoder
We design the encoder and decoder similar to the generator and discriminator of DCGAN~\citep{dcgan}.
The details of the architecture are at \autoref{tab:architectures}.
We use learning rate 1e-3, batch size 100, and Adam optimizer for training.
We use Bernoulli loss as the reconstruction loss for Breakout experiments and negative log-likelihood loss as the reconstruction loss for CUB, Food101, and Oxford102 experiments.
We use the validation set for early stopping and report the negative log-likelihood on the test set with 50 importance weighted samples.

\begin{table}[h!]
\centering
\caption{The architectures of encoder and decoder used in the density estimation experiments. $n_c$ is the number of channels of the image, $n_d$ is the latent dimension. $n_a$ is a coefficient that depends on the VAE, i.e., $n_a$ is 2, 2, 1.5, 1.5 for $\mathcal{E}$-VAE, $\mathcal{L}$-VAE, $\mathcal{P}$-VAE, and GM-VAE, respectively. }
\label{tab:architectures}
\vskip 0.1in
\begin{tabular}{l l}
\toprule
Encoder \\
\midrule
Layer & Size \\
\midrule
Input & $64 \times 64 \times n_c$ \\
Convolution2D & $32 \times 32 \times 32$ \\
LeakyReLU \\
Convolution2D & $16 \times 16 \times 64$ \\
LeakyReLU \\
Convolution2D & $8 \times 8 \times 128$ \\
LeakyReLU \\
Convolution2D & $4 \times 4 \times 256$ \\
LeakyReLU \\
Linear & $n_a \cdot n_d$ \\
\bottomrule
\end{tabular}
\quad
\begin{tabular}{l l}
\toprule
Decoder \\
\midrule
Layer & Size \\
\midrule
Input & $1 \times 1 \times n_d$ \\
TransposedConvolution2D & $4 \times 4 \times 256$ \\
ReLU \\
TransposedConvolution2D & $8 \times 8 \times 128$ \\
ReLU \\
TransposedConvolution2D & $16 \times 16 \times 64$ \\
ReLU \\
TransposedConvolution2D & $32 \times 32 \times 32$ \\
ReLU \\
TransposedConvolution2D & $64 \times 64 \times n_c$ \\
\bottomrule
\end{tabular}
\end{table}

% Hyper-parameters
% learning rate, batch size, epochs, optimizer, recon loss

% Evaluation
% Validation set, importance weighted sampling

\subsection{Model-based RL}
\label{apx:rl_details}

We use the official TensorFlow implementation from Dreamerv2\footnote{\href{https://github.com/danijar/dreamerv2}{https://github.com/danijar/dreamerv2}} to reproduce the baseline results, i.e., with Euclidean and discrete latent space.
For the hyperbolic latent space results, we apply GM-VAE by replacing the latent space of $z_t$ with the Gaussian manifold and two components in RSSM, the representation model $q_\theta(z_t | h_t, x_t)$ and transition predictor $p_\phi(z_t | h_t)$, with PGM normal.
The hyperparameters are set to be the same as suggested in the original paper, except for the training environment steps being 50M for Freeway and 100M for the others as we observe converging scores.
% use the official tensorflow implementation from Dreamerv2.
% the hyper-parameters are set to be same.

\section{Results of Non-Product Latent Space}
\label{apx:non_product}

Previous hyperbolic VAEs are implemented with a hyperbolic space, not the product of the hyperbolic spaces.
We run the non-product hyperbolic VAEs in the density estimation of image datasets.
\autoref{tab:density_estimation_results_nonproduct} reveals that the non-product hyperbolic VAEs fail in most of the settings and the product hyperbolic space makes the hyperbolic VAEs much more stable.

\begin{table*}[h!]
    \centering
    \vskip -0.1in
    \caption{Density estimation results of non-product hyperbolic VAEs. $d$ denotes the latent dimension. N/A in the log-likelihood indicates that the results are not available due to the failure of all runs.}
    \vskip 0.1in
    %\resizebox{\textwidth}{!}{%
    \begin{tabular}{l l r r}
        \toprule
        & $d$ & $\mathcal{L}$-VAE & $\mathcal{P}$-VAE \\
         \midrule
         \multirow{3}{*}{Breakout} 
         &2&$124.24_{\pm1.66}$&$266.86_{\pm6.01}$\\
         &4&$66.20_{\pm0.14}$&N/A\\
         &8&$44.76_{\pm0.48}$&N/A\\
         \midrule
         \multirow{3}{*}{CUB} 
         &50&N/A&N/A\\
         &60&N/A&N/A\\
         &70&N/A&N/A\\
         \midrule
         \multirow{3}{*}{Oxford102}
         &50&N/A&N/A\\
         &60&N/A&N/A\\
         &70&N/A&N/A\\
         \bottomrule
    \end{tabular}%
    %}
    \label{tab:density_estimation_results_nonproduct}
\end{table*}

%\section{Details of density estimation}

% \input{appendices/06_NumericalStability.tex}
\section{Model-Based RL}
\subsection{Learning curves}

\begin{figure}[h!]
    \centering
    \includegraphics[width=.7\linewidth]{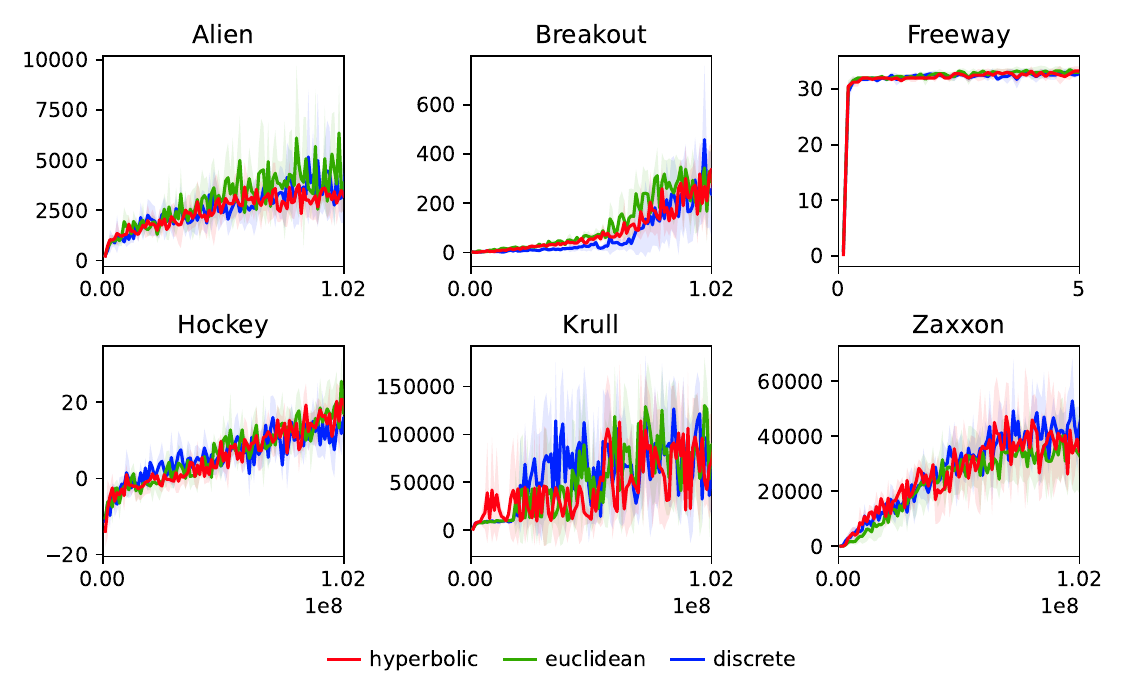}
\end{figure}

\subsection{Latent space analysis}
\label{apx:rl_analysis}

% We compute hierarchy preservation.
We conduct an analysis of the latent space of the agents learned to play Atari2600 Breakout.
The purpose of the analysis is to measure how the latent spaces well-preserve the implicit hierarchy in the trajectory of the agents.
To analyze the hyperbolic latent space, we need two isometries: the isometry between the Gaussian manifold and the Poincar\'e disk model and the translation of the Poincar\'e disk model.

% First, isometries are (halfplane <-> disk, translation in disk)
We first propose an isometry between the Gaussian manifold and the Poincar\'e disk model $T_{\mathcal{P}_c \rightarrow \mathcal{U}_c}: \mathcal{P}_c \rightarrow \mathcal{U}_c$:
\begin{equation*}
    T_{\mathcal{P}_c \rightarrow \mathcal{G}_c}(x, y) = \left( \frac{-2y}{(\sqrt{c}x - 1)^2 + y^2c}, \frac{1 - (x^2 + y^2)c}{(\sqrt{c}x - 1)^2 + y^2c} \right), 
\end{equation*}
and the inverse is:
\begin{equation*}
     T^{-1}_{\mathcal{P}_c \rightarrow \mathcal{G}_c}(x, y) \left( \frac{\sqrt{c}x^2 + (y^2 - 1) / \sqrt{c}}{cx^2 + (y + 1)^2}, \frac{-2x}{cx^2 + (y + 1)^2} \right).
\end{equation*}

The translation of the Poincar\'e disk model can be derived using complex numbers. 
Let $z = x + yi \in \mathbb{C}$ and $(x, y) \in \mathcal{P}_1$ and $z_0 \in \mathbb{C}$ be the pivot point. 
Then the isometry that moves $z_0$ to the origin is defined as $T(z): \mathcal{P}_1 \rightarrow \mathcal{P}_1 := \frac{z - z_0}{1 - \bar{z_0}z}$.
Note that the translation of Euclidean space is $z-z_0$.

% Second, we iterate over the points and translate them to the origin (and the others with the same amount)
After transforming the latents on the Gaussian manifold to the Poincar\'e disk model and using the translation, we can measure how the latents well-captures the hierarchical structure of data.
We first pick a latent and then translate all the latents by setting the selected latent as the pivot point.
We then measure the Pearson correlation between the cumulative reward of the latents and the norm.

We repeat this process for all the latents and compute the maximum of the correlations.
We use the latents obtained from the agents which recorded at least 250 for long enough trajectories.
We obtain a correlation coefficient of 0.46 from the hyperbolic latent space, whereas the correlation coefficient of the Euclidean latent space is 0.40, showing the hyperbolic space better captures the hierarchy along the increasing norm.

% Last, we compute the correlation between the norm and cumulative reward and report the best value.

% Other latent visualization?

%%%%%%%%%%%%%%%%%%%%%%%%%%%%%%%%%%%%%%%%%%%%%%%%%%%%%%%%%%%%

\end{document}